%% file: acl_latex.tex
\pdfoutput=1

\documentclass[11pt]{article}

\usepackage[]{acl}

\usepackage{times}
\usepackage{latexsym}

\usepackage[T1]{fontenc}

\usepackage[utf8]{inputenc}

\usepackage{microtype}

\usepackage{inconsolata}

\usepackage{longtable}
\usepackage{tabu}
\usepackage{arydshln}

\usepackage{xspace}
\usepackage{graphicx}
\usepackage{wrapfig}
\usepackage{CJK,algorithm,algorithmic,amssymb,amsmath,array,epsfig,graphics,float,subfigure,verbatim,epstopdf}
\usepackage{enumitem}
\usepackage{color,soul}
\usepackage{hhline}
\usepackage{multirow}
\usepackage{multicol}
\usepackage{xcolor}
\usepackage{hyperref}
\usepackage{cleveref}

\usepackage{xurl}

\usepackage{tabularx}
\usepackage{bm}
\usepackage{seqsplit}
\usepackage{stmaryrd}
\usepackage{booktabs}
\usepackage{xparse}

\usepackage{fixltx2e}

\usepackage{comment} 
\NewDocumentCommand{\heng}
{ mO{} }{\textcolor{red}{\textsuperscript{\textit{Heng}}\textsf{\textbf{\small[#1]}}}}
\NewDocumentCommand{\qingyun}
{ mO{} }{\textcolor{cyan}{\textsuperscript{\textit{qingyun}}\textsf{\textbf{\small[#1]}}}}
\NewDocumentCommand{\zixuan}
{ mO{} }{\textcolor{orange}{\textsuperscript{\textit{zixuan}}\textsf{\textbf{\small[#1]}}}}

\title{Chem-FINESE: Validating Fine-Grained Few-shot Entity Extraction through Text Reconstruction
}

\author{
Qingyun Wang, \ Zixuan Zhang, \ Hongxiang Li, \ Xuan Liu, \\ \  \textbf{Jiawei Han}, \ \textbf{Huimin Zhao}, \ \textbf{Heng Ji}\\   
 University of Illinois at Urbana-Champaign \\
\texttt{\fontfamily{pcr}\selectfont\{qingyun4,zixuan11,hanj,zhao5,hengji\}@illinois.edu}\\
}

\definecolor{bleudefrance}{rgb}{0.19, 0.55, 0.91}
\definecolor{auburn}{rgb}{0.43, 0.21, 0.1}
\definecolor{ao}{rgb}{0.0, 0.5, 0.0}
\definecolor{purp}{RGB}{157,78,221}

\definecolor{myorange}{HTML}{E07234}
\definecolor{myred}{HTML}{D40c27}
\definecolor{mygreen}{HTML}{81A355}

\begin{document}
\maketitle

\input{0abstract}
\input{1introduction}

\input{2task}

\input{3method}

\input{4data}

\input{5experiment}

\input{6analysis}

\input{7related}
\input{8conclusion}
\input{9limitations}

\section*{Acknowledgement}
This work is supported by the Molecule Maker Lab Institute: an AI research institute program supported by NSF under award No. 2019897 , and by DOE Center for Advanced Bioenergy and Bioproducts Innovation U.S. Department of Energy, Office of Science, Office of Biological and Environmental Research under Award Number DESC0018420. The views and conclusions contained herein are those of the authors and should not be interpreted as necessarily representing the official policies, either expressed or implied of, the National Science Foundation, the U.S. Department of Energy, and the U.S. Government. The U.S. Government is authorized to reproduce and distribute reprints for governmental purposes notwithstanding any copyright annotation therein.
\bibliography{anthology,custom}

\appendix

\input{appendix}


\end{document}

%% file: 0abstract.tex
\begin{abstract}
Fine-grained few-shot entity extraction in the chemical domain faces two unique challenges. First, compared with entity extraction tasks in the general domain, sentences from chemical papers usually contain more entities. Moreover, entity extraction models usually have difficulty extracting entities of long-tailed types. In this paper, we propose Chem-FINESE, a novel sequence-to-sequence (seq2seq) based few-shot entity extraction approach, to address these two challenges. Our Chem-FINESE has two components: a seq2seq entity extractor to extract named entities from the input sentence and a seq2seq self-validation module to reconstruct the original input sentence from extracted entities. Inspired by the fact that a good entity extraction system needs to extract entities faithfully, our new self-validation module leverages entity extraction results to reconstruct the original input sentence. Besides, we design a new contrastive loss to reduce excessive copying during the extraction process. Finally, we release ChemNER+, a new fine-grained chemical entity extraction dataset that is annotated by domain experts with the ChemNER schema. Experiments in few-shot settings with both ChemNER+ and CHEMET datasets show that our newly proposed framework has contributed up to 8.26\% and 6.84\% absolute F1-score gains respectively\footnote{The programs, data, and resources are publicly available for research purposes at: \url{https://github.com/EagleW/Chem-FINESE}.}. 
\end{abstract}

%% file: 1introduction.tex
\section{Introduction}

Millions of scientific papers are published annually\footnote{\url{https://esperr.github.io/pubmed-by-year/about.html}}, resulting in an information overload
~\cite{van2014global,landhuis2016scientific}. 
Due to such an explosion of research directions, it is impossible for scientists to fully explore the landscape due to the limited reading ability of humans.
Therefore, information extraction, especially entity extraction of fine-grained scientific entity types, becomes a crucial step to automatically catch up with the newest research findings in the chemical domain.

\begin{figure}[t]
\centering
\includegraphics[width=\linewidth]{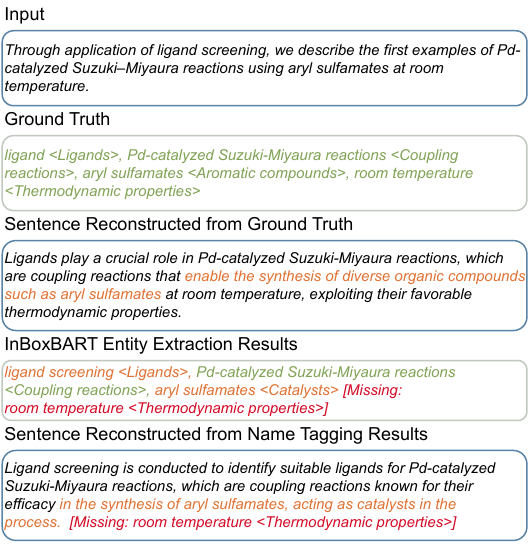}
\caption{\label{fig:exp} Comparison of sentence reconstruction results from ground truth and InBoXBART~\cite{parmar-etal-2022-boxbart}. We highlight \textcolor{mygreen}{Complete Correct}, \textcolor{myred}{\textit{Missed Entity}}, and \textcolor{myorange}{\textit{Partially Correct Prediction}} with different color. 
}
\end{figure}

Despite such a pressing need, fine-grained entity extraction in the chemical domain presents three distinctive and non-trivial challenges. 
First, there are very few publicly available benchmarks with high-quality annotations on fine-grained chemical entity types. 
For example, ChemNER~\cite{wang-etal-2021-chemner} developed the first fine-grained chemistry entity extraction dataset.
However, their dataset is not released publicly. 
To address this issue, we collaborate with domain experts to annotate ChemNER+, a new chemical entity extraction dataset based on the ChemNER ontology. Besides, we construct another new fine-grained entity extraction dataset based on an existing entity typing dataset CHEMET~\cite{checnkai2021}.

\begin{figure}[!hbt]
\centering
\includegraphics[width=\linewidth]{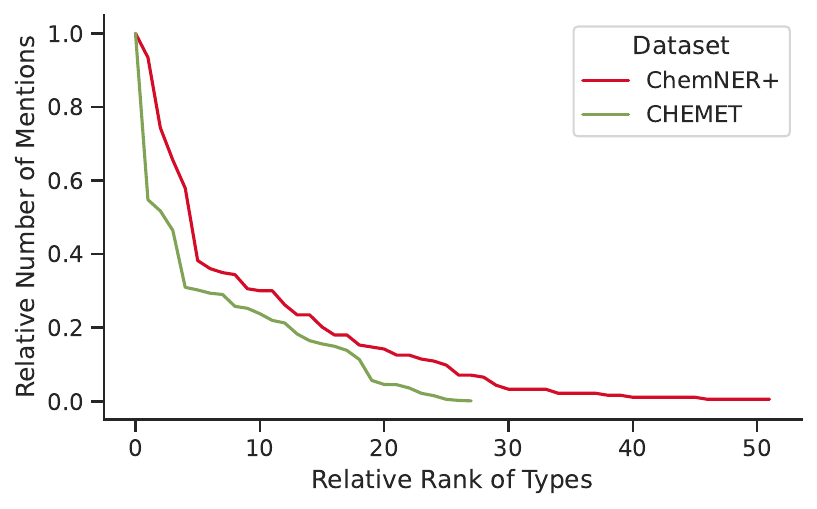}
\caption{Type distributions for the training sets of ChemNER+ and CHEMET datasets. The Y-axis represents the number of mentions normalized by the mentions of the most frequent type.  The X-axis represents the rank of types.}
\label{img:longtail}
\end{figure}

In addition, current entity extraction systems in few-shot settings face two main problems: \textit{missing mentions} and \textit{incorrect long-tail predictions}.
One primary reason for missing mentions is that the sentences in scientific papers typically cover more entities than sentences in the general domain. 
For example, there are 3.1 entities per sentence in our ChemNER+ dataset, which is much higher than the 1.5 entities in the general domain dataset CONLL2003~\cite{tjong-kim-sang-de-meulder-2003-introduction}. As a result, it is more difficult for entity extraction models to cover all mentions in the input sentences.
As shown in Figure~\ref{fig:exp}, since the input has already included four chemical entities, InBoXBART model~\cite{parmar-etal-2022-boxbart} completely misses the entity \textit{``room temperature''}. 

Furthermore, entity distributions in the chemical domain are highly imbalanced. 
As shown in Figure~\ref{img:longtail}, we observe that the entity type distributions of ChemNER+ and CHEMET exhibit similar long-tail patterns. 
In few-shot settings, entities with long-tail types are extremely difficult to extract due to insufficient training examples. 
For example, as shown in Figure~\ref{fig:exp}, InBoXBART mistakenly predicts the entity \textit{``aryl sulfamates''} as \textit{catalyst}, because its type has a frequency forty times lower than the predicted type (i.e., 4 vs 136). 
Moreover, the diverse representation nature of chemical entities---such as trade names, trivial names, and semi-systematic names (e.g., THF, iPrMgCl, 8-phenyl ring)---makes it even harder for models to generalize on these long-tail entities. 

To address these challenges, we propose a novel \textbf{Chem}ical \textbf{FIN}e-grained \textbf{E}ntity extraction with \textbf{SE}lf-validation (Chem-FINESE). Specifically, our Chem-FINESE has two parts: a seq2seq entity extractor to extract named entities from the input sentence and a seq2seq self-validation module to reconstruct the original input sentence based on the extracted entities. First, we employ a seq2seq model to extract entities from the input sentence, since it does not require any task-specific component and explicit negative training examples~\cite{giorgi-etal-2022-sequence}. We generate the entity extraction results as a concatenation of pairs, each consisting of an entity mention and its corresponding type, as shown in Figure~\ref{fig:exp}. 

One critical issue for seq2seq entity extraction is that the language model tends to miss important entities or excessively copy original input. For example, the seq2seq entity extraction results missed the type \textit{thermodynamic properties} and generated \textit{``ligand screening''} in Figure~\ref{fig:exp}. However, the goal of information extraction is to provide factual information and knowledge comprehensively. In other words, \textit{if the model extracts knowledge precisely, readers should be able to faithfully reconstruct the original sentence using the extraction results}. Inspired by such a goal, to evaluate whether the seq2seq entity extractor has faithfully extracted important information, we propose a novel seq2seq self-validation module to reconstruct the original sentences based on entity extraction results. As shown in Figure~\ref{fig:exp}, the sentence reconstructed from the ground truth is closer to the original input than the sentence reconstructed from entity extraction results, which misses the reaction condition and introduces additional information that treated the \textit{``aryl sulfamates''} as \textit{catalysts}. Additionally, we introduce a new entity decoder contrastive loss to control the mention spans. We treat text spans containing entity mentions as hard negatives. For instance, given the ground truth entity \textit{``aryl sulfamates''}, we will treat \textit{``aryl sulfamates at room temperature''} as a hard negative.

Our extensive experiments demonstrate that our proposed framework significantly outperforms our baseline model by up to 8.26\% and 6.84\% absolute F1-score gains on ChemNER+ and CHEMET datasets respectively. Our analysis also shows that Chem-FINESE can effectively learn to select correct mentions and improve long-tail entity type performance. To evaluate the generalization ability of our proposed method, we also evaluate our framework on CrossNER~\cite{liu2021crossner}, which is based on Wikipedia. Our Chem-FINESE still outperforms other baselines in all five domains. 

Our contributions are threefold: 

\begin{enumerate}
\item We propose two few-shot chemical fine-grained entity extraction datasets, based on human-annotated ChemNER+ and CHEMET.
\item We propose a new framework to address the mention coverage and long-tailed entity type problems in chemical fine-grained entity extraction tasks through a novel self-validation module and a new entity extractor decoder contrastive objective. Our model does not require any external knowledge or domain adaptive pretraining. 
\item Our extensive experiments on both chemical few-shot fine-grained datasets and the CrossNER dataset justify the superiority of our Chem-FINESE model.
\end{enumerate}

%% file: 2task.tex
\section{Task Formulation}
\label{sec:task}
Following \citet{giorgi-etal-2022-sequence}, we formulate entity extraction as a sequence-to-sequence (seq2seq) generation task by taking a source document $\mathcal{S}$ as input. The model generates output $\mathcal{Y}$, a text consisting of a concatenation of $n$ fine-grained chemical entities $E_1, E_2, ..., E_n$. Each mention $E_i$ includes the mention $\mu_i$ in the source document $\mathcal{S}$ and its entity type $\rho_i\in\mathcal{P}$, where $\mathcal{P}$ is a set containing all entity types. 
Specifically, we propose the following output linearization schema: given the input $\mathcal{S}$, the output is $\mathcal{Y}=\mu_1 <\rho_1>,\mu_2 <\rho_2>,...,\mu_n <\rho_n>$. We further illustrated this with an example:

\noindent
{\small
\(\mathcal{S}\): Through application of \textcolor{mygreen}{ligand} screening, we describe the first examples of \textcolor{mygreen}{Pd-catalyzed Suzuki–Miyaura reactions} using \textcolor{mygreen}{aryl sulfamates} at \textcolor{mygreen}{room temperature}.
}\\
\noindent
{\small
\(\mathcal{Y}\): \textcolor{mygreen}{ligand} <Ligands>, \textcolor{mygreen}{Pd-catalyzed Suzuki–Miyaura reactions} <Coupling reactions>, \textcolor{mygreen}{aryl sulfamates} <Aromatic compounds>, \textcolor{mygreen}{room temperature} <Thermodynamic properties>
}\\

\begin{figure}[!bt]
\centering
\includegraphics[width=\linewidth]{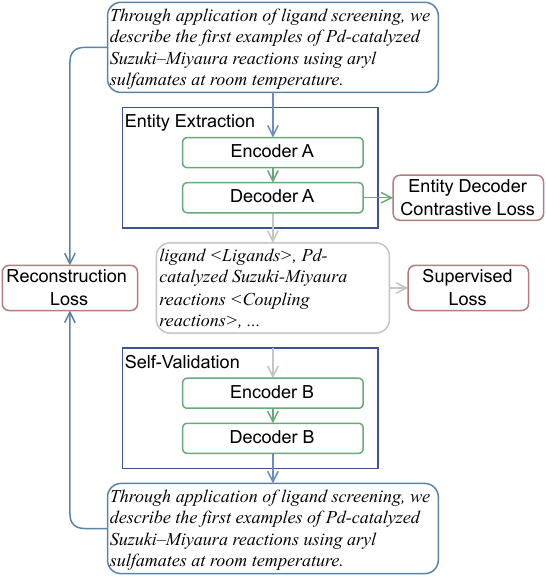}
\caption{Architecture overview. We use the example in Figure~\ref{fig:exp} as a walking-through example.
}
\label{img:overview}
\end{figure}

%% file: 3method.tex
\section{Method}

\input{3.1overview}

\input{3.2ner}
\input{3.3self}
\input{3.4cl}
\input{3.5obj}

%% file: 3.1overview.tex
\subsection{Model Architecture}

The overall framework is illustrated in Figure \ref{img:overview}. Given the source document $S$, we first use a seq2seq model to extract fine-grained chemical entities. Then, we propose a new \textit{self-validation module} to reconstruct the original input based on entity extraction results.  Finally, we introduce a new \textit{entity decoder contrastive loss} to reduce excessive copying. The entire model is trained with a combination of the supervised loss, the reconstruction loss, and the entity decoder contrastive loss.

%% file: 3.2ner.tex
\subsection{Entity Extraction Module}

Our entity extraction module follows a seq2seq setup~\cite{yan-etal-2021-unified-generative,giorgi-etal-2022-sequence}. Formally, we use the state-of-the-art coarse-grained chemical entity extractor InBoXBART~\cite{parmar-etal-2022-boxbart} as the backbone. We model the conditional probability of extracting entities from source sequence $\mathcal{S}$ as
\begin{equation}
    p(\mathcal{Y}|\mathcal{S}) = \prod_{t=1}^{T} p(y_t|\mathcal{S},y_{<t}),
\end{equation}
where the output $\mathcal{Y}$ has a length of $T$, and $y_t$ is the predicted token at time $t$ in the output $\mathcal{Y}$. 

We supervise the entity extraction using the standard cross-entropy loss:
\begin{equation}
\mathcal{L}_{\mathrm{gen}} = - \sum_{t=1}^{T} \log p(y_t|\mathcal{S},y_{<t}).
\end{equation}

%% file: 3.3self.tex
\subsection{Self-validation Module}
Since a good information extraction system needs to extract entities faithfully, we propose a self-validation module to reconstruct the original sentence from the extracted entities to check whether the model overlooks any entities. Different from previous dual learning architectures~\cite{cyclener}, which use dual cycles or reinforcement learning to provide feedback, we use Gumbel-softmax (GS) estimator~\cite{jang2017categorical} to avoid the non-differentiable issue in explicit decoding. Specifically, based on InBoXBART~\cite{parmar-etal-2022-boxbart}, we first pretrain a seq2seq self-validation module that takes in the entity extraction results $\mathcal{Y}$ and generates a reconstructed sentence $\mathcal{\hat{S}}$. 
We use our training set to pretrain the self-validation module.
We fix the weight of the self-validation module after pretraining. In the training stage, the input embedding $\mathbf{H}_t$ of the self-validation module is given by:
\begin{align}
    \mathbf{H}_t = \mathsf{GS}\left(p\left(y_t|\mathcal{S},y_{<t}\right)\right) \cdot \mathbf{E}_v,
\end{align}
where $\mathbf{E}_v$ is the vocabulary embedding matrix and $\mathsf{GS}$ is the Gumbel-softmax estimator. The total input embeddings for the self-reconstruction model is $\mathbf{H}=[\mathbf{H}_1;\mathbf{H}_2;...;\mathbf{H}_T]$.

The reconstruction loss is:
\begin{equation}
\mathcal{L}_{\mathrm{recon}} = -\sum_{\hat{t}=1}^{\hat{T}} \log p(\hat{s}_{\hat{t}}|\mathbf{H},\hat{s}_{<\hat{t}}),
\end{equation}
where the reconstructed sentence $\hat{\mathcal{S}}$ has a length of $\hat{T}$, and  $\hat{s}_{\hat{t}}$ is the predicted token at time $\hat{t}$ in $\hat{\mathcal{S}}$.

%% file: 3.4cl.tex
\subsection{Contrastive Entity Decoding Module}

Entity extraction datasets in the scientific domain usually contain more entities for each sentence. 
From the initial experiments, we found that the entity extraction module tends to generate incorrect mentions by associating it with unrelated contexts
to help the reconstruction of the self-validation module. 
For example, given the example in Figure~\ref{fig:exp}, the baseline model generates \textit{``ligand screening''} instead of \textit{``ligand''}.
Therefore, we introduce a new decoding contrastive loss inspired by~\citet{wang-etal-2023-multimedia} to suppress excessive copying. 
We construct negative samples by combining mentions with surrounding unrelated contexts. 
For example, we will consider \textit{``ligand screening, we describe the first examples''} as a negative of entity \textit{``ligand''}.
We treat the original mention type pairs as the ground truth and maximize their probability with InfoNCE loss~\cite{oord2018representation}: 

\begin{align}
\begin{split}
    \mathcal{L}_{\mathrm{cl}} &= \frac{\exp{\left(x^+/\tau\right)}}{\sum_i \exp{\left(x^-_i/\tau\right)} +\exp{\left(x^+/\tau\right)} }, \\
    x^+&=\sigma(\mathrm{Avg}(\mathbf{W}_x\mathbf{\bar{H}}^++\mathbf{b}_x)),\\
    x^-_i&=\sigma(\mathrm{Avg}(\mathbf{W}_x\mathbf{\bar{H}}^-_i+\mathbf{b}_x)),\\
\end{split}
\end{align}
where $\mathbf{\bar{H}}^+$ and $\mathbf{\bar{H}}_i^-$ are decoder hidden states from the positive and $i$-th negative samples, $\mathbf{W}_x$ is a learnable parameter, $\tau$ is the temperature, and $\mathrm{Avg}(*)$ denotes the average pooling function.

%% file: 3.5obj.tex
\subsection{Training Objective}

We jointly optimize the cross-entropy loss, reconstruction loss, and entity decoder contrastive loss: 
\begin{equation}
  \mathcal{L} = \mathcal{L}_{\mathrm{gen}} + \alpha\mathcal{L}_{\mathrm{recon}} + \beta \mathcal{L}_{\mathrm{cl}},
\end{equation}
where $\alpha,\beta$ are hyperparameters that control the weights of the reconstruction loss and contrastive loss respectively.

%% file: 4data.tex
 \begin{table}[!htb]
\centering
\small
\begin{tabularx}{\linewidth}{>{\hsize=1.4\hsize}X>{\arraybackslash\hsize=0.8\hsize}X>{\centering\arraybackslash\hsize=0.8\hsize}X>{\centering\arraybackslash\hsize=1\hsize}X>{\centering\arraybackslash\hsize=1\hsize}X}
\toprule
\textbf{Dataset}&\textbf{Split}   & \textbf{\#Pair}& $\mathbf{\overline{\#Token}}$ & $\mathbf{\overline{\#Entity}}$ \\
\midrule
                &Train       & 542  & 32.9 & 3.10 \\
ChemNER+         &Valid      & 100  & 39.9 & 4.57\\
                &Test        & 100  & 39.4 & 4.61 \\
\hdashline
                &Train       & 6,561  & 37.8 & 1.57 \\
CHEMET          &Valid      & 520    & 31.6 & 2.15 \\
                &Test        & 663    & 36.6 & 1.95 \\
\bottomrule
\end{tabularx}
\caption{Statistics of our dataset. $\mathbf{\overline{\#Token}}$ denotes average number of words per sentence. $\mathbf{\overline{\#Entity}}$ denotes average number of entities per sentence. \label{tab:stat} }
\end{table}

\section{Benchmark Dataset}

\subsection{Dataset Creation}
\paragraph{ChemNER+ Dataset.}
Since the annotation of ChemNER dataset is not fully available online, we decide to create our own dataset, ChemNER+, based on available sentences from ChemNER~\cite{wang-etal-2021-chemner} dataset. Following the schema of ChemNER, we ask two Chemistry Ph.D. students to annotate a new dataset, covering 59 fine-grained chemistry types with 742 sentences\footnote{Human annotation details are in Appendix~\ref{app:human}.}.

\paragraph{CHEMET Dataset.}
We construct a new fine-grained entity extraction dataset based on CHEMET~\cite{checnkai2021}. For any entity in the training set that overlaps with the validation and testing sets, we replace its multi-labels with the most frequent types that appear in the validation and testing sets. For other entities, we replace the remaining types with their most frequent types that appeared in the training set. We merge the entity types with the same subcategory name in CHEMET~\cite{checnkai2021}. The final dataset consists of 30 fine-grained organic chemical types.

Table~\ref{tab:stat} shows the detailed data statistics.

\input{4.1setup}

%% file: 4.1setup.tex
\begin{table}[!htb]
\centering
\small
\begin{tabularx}{\linewidth}{>{\hsize=2.15\hsize}X>{\centering\arraybackslash\hsize=0.77\hsize}X>{\centering\arraybackslash\hsize=0.77\hsize}X>{\centering\arraybackslash\hsize=0.77\hsize}X>{\centering\arraybackslash\hsize=0.77\hsize}X>{\centering\arraybackslash\hsize=0.77\hsize}X}
\toprule
 \textbf{$k$-shot} &\textbf{6} &\textbf{9}&\textbf{12} &\textbf{15}&\textbf{18} \\
 \midrule
RoBERTa              & 8.09 & 7.98& 8.00 & 16.22 & 7.94\\ 
PubMedBERT           & 5.48 & 5.12& 5.77 & 5.46 & 5.88\\ 
ScholarBERT          & 23.96  &29.82 & 27.65 &31.48  & 32.76\\ 
\hdashline
NNShot	             &0.99	&1.43	&2.39	&1.61	&2.45\\
StructShot	         &0.86	&1.32	&2.27	&1.62	&2.47\\
\hdashline
InBoXBART            & 26.23  &27.89 &28.83  & 33.64 & 30.39\\ 
+ Valid              & 32.40 &31.13 & 33.64 & 35.31 & 36.44\\ 
+ Valid + CL         & \textbf{33.11} &\textbf{32.75} & \textbf{34.75} & \textbf{37.89} & \textbf{38.65}\\ 
\bottomrule
\end{tabularx}
\caption{micro-F1 (\%) scores for ChemNER+ with few-shot settings. \textit{Valid} is a model with a self-validation module. \textit{CL} is a model with a decoder contrastive loss.
}
\label{tab:few_ChemNER+}
\end{table}

\begin{table}[!htb]

\centering
\small
\begin{tabularx}{\linewidth}{>{\hsize=2.15\hsize}X>{\centering\arraybackslash\hsize=0.77\hsize}X>{\centering\arraybackslash\hsize=0.77\hsize}X>{\centering\arraybackslash\hsize=0.77\hsize}X>{\centering\arraybackslash\hsize=0.77\hsize}X>{\centering\arraybackslash\hsize=0.77\hsize}X}
\toprule
 \textbf{$k$-shot} &\textbf{6} &\textbf{9}&\textbf{12} &\textbf{15}&\textbf{18} \\
 \midrule
RoBERTa              &  4.91 &  4.16& 4.79 & 4.83 & 4.81\\ 
PubMedBERT           &  4.07 &  4.67 & 3.87 & 4.47 & 3.96\\ 
ScholarBERT          & 17.00 & 33.63& 29.65  &29.72  & 32.52\\ 
\hdashline
NNShot	             &4.23	&4.03	&4.14	&5.27	&4.76\\
StructShot	         &4.15	&4.00	&4.19	&5.21	&4.79\\
\hdashline
InBoXBART            & 29.93 &29.57 &31.76  &36.16  & 37.52\\ 
+ Valid              & 32.74 &34.09 &33.30  & \textbf{40.81} & 38.37\\ 
+ Valid + CL    &\textbf{33.81}  & \textbf{36.41}& \textbf{36.11} & 40.52 &\textbf{39.94} \\ 
\bottomrule
\end{tabularx}
\caption{micro-F1 (\%) scores for CHEMET with few-shot settings.
}
\label{tab:few_chemet}
\end{table}

\subsection{Few-shot Setup}
For each dataset, we randomly sample a subset based on the frequency of each type class. Specifically, given a dataset, we first set the number of maximum entity mentions $k$ for the most frequent entity type in the dataset. We then randomly sample other types and ensure that the distribution of each type remains the same as in the original dataset. We choose the values $6, 9, 12, 15, 18$ as the potential maximum entity mentions for $k$. The ChemNER+ and CHEMET few-shot datasets contain 52 and 28 types respectively.

%% file: 5experiment.tex
\section{Experiments}

\input{5.1baseline}
\input{5.2automatic}

%% file: 5.1baseline.tex
\subsection{Baselines}

We compare our model with \textbf{(1) state-of-the-art pretrained encoder-based models} including RoBERTa~\cite{liu2019roberta} and models with domain adaptive training, such as PubMedBERT~\cite{pubmedbert} and ScholarBERT~\cite{hong-etal-2023-diminishing}. We then compare our model with the \textbf{(2) few-shot baselines}, including NNShot and StructShot~\cite{yang-katiyar-2020-simple} based on RoBERTa-base. Since we use \textbf{InBoXBART}~\cite{parmar-etal-2022-boxbart} as our backbone, we also include \textbf{(3) baselines for ablation}. The hyperparameters, training and evaluation details are presented in Appendix~\ref{app:hyper}. 

%% file: 5.2automatic.tex
\subsection{Overall Performance}

Tables~\ref{tab:few_ChemNER+}, \ref{tab:few_chemet} show that our models outperform baselines for few-shot settings by a large margin. Compared to the best pretrained encoder-based ScholarBERT, pretrained on 221B tokens of scientific documents, seq2seq models generally achieve higher performance in low-resource settings with fewer parameters, as shown in Table~\ref{tab:para}. We also observe that both NNshot and StructShot perform worse than their original baseline. At a closer look, we find that both methods miss many entities and mislabel unrelated phrases as entities. The primary reasons for this are twofold: first, the chemical domain's entity mentions are more diverse and may only appear in the testing set; second, there are significantly more potential entity types than in traditional entity extraction tasks. Therefore, the two baselines cannot effectively utilize the nearest neighbor information and perform worse than our proposed methods.
These results demonstrate that seq2seq models have a better generalization ability in few-shot settings. 

\begin{table}[!htb]
\centering
\small
\begin{tabularx}{\linewidth}{>{\hsize=2.15\hsize}X>{\centering\arraybackslash\hsize=0.77\hsize}X>{\centering\arraybackslash\hsize=0.77\hsize}X>{\centering\arraybackslash\hsize=0.77\hsize}X>{\centering\arraybackslash\hsize=0.77\hsize}X>{\centering\arraybackslash\hsize=0.77\hsize}X
}
\toprule
 \textbf{$k$-shot} &\textbf{6} &\textbf{9}&\textbf{12} &\textbf{15}&\textbf{18} \\
 \midrule
InBoXBART           & 36.96  & 38.22 & 38.34 & 47.91 & 42.84\\ 
+ Valid             & 45.07  & \textbf{45.28} & 41.56 & 48.15& 46.15\\ 
+ Valid + CL        & \textbf{45.58}  & 44.03 & \textbf{45.25} & \textbf{51.68} & \textbf{47.88}\\ 
\bottomrule
\end{tabularx}
\caption{Mention micro-F1 (\%) scores for ChemNER+ with few-shot settings. 
}
\label{tab:few_ChemNER+_mention}
\end{table}

\begin{table}[!htb]

\centering
\small
\begin{tabularx}{\linewidth}{>{\hsize=2.15\hsize}X>{\centering\arraybackslash\hsize=0.77\hsize}X>{\centering\arraybackslash\hsize=0.77\hsize}X>{\centering\arraybackslash\hsize=0.77\hsize}X>{\centering\arraybackslash\hsize=0.77\hsize}X>{\centering\arraybackslash\hsize=0.77\hsize}X
}
\toprule
 \textbf{$k$-shot} &\textbf{6} &\textbf{9}&\textbf{12} &\textbf{15}&\textbf{18} \\
 \midrule
InBoXBART           & 46.74  & 42.07& 44.32 & 47.58 & 52.90\\ 
+ Valid             & 47.87  & 46.01& 44.18 & 50.55 & 50.50\\ 
+ Valid + CL        & \textbf{48.96}  & \textbf{49.83}& \textbf{47.03} & \textbf{50.61} & \textbf{54.10}\\ 
\bottomrule
\end{tabularx}
\caption{Mention micro-F1 (\%) scores for CHEMET with few-shot settings.
}
\label{tab:few_chemet_mention}
\end{table}

Additionally, the self-validation variants significantly outperform the baseline InBoXBART, showing the benefit of the self-validation module in capturing mentions. Moreover, our self-validation module can effectively enhance the performance of the entity extraction module in extremely low-resource settings. In 6-shot scenarios for both ChemNER+ and CHEMET datasets, our model achieves impressive performance compared to ScholarBERT, which further verifies the effectiveness of the self-validation module. Finally, adding decoder contrastive loss helps the model perform significantly better in Table~\ref{tab:few_ChemNER+}, suggesting that contrastive learning further helps the mention extraction quality by reducing excessive copying. Interestingly, we observe that decoder contrastive learning improves less in Table~\ref{tab:few_chemet} than in Table~\ref{tab:few_ChemNER+}, because the CHEMET contains fewer entities per sentence compared to the ChemNER+.


\begin{figure}[!hbt]
\centering
\includegraphics[width=\linewidth]{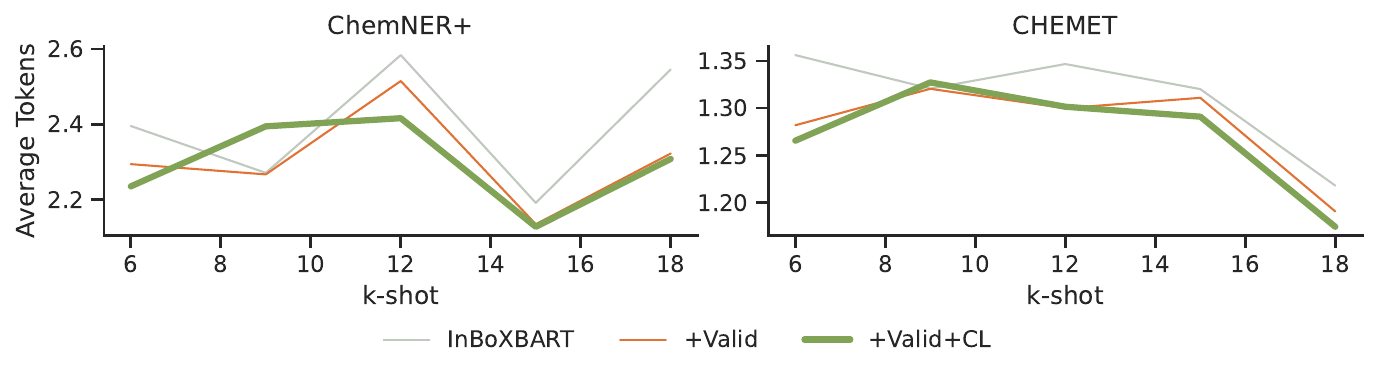}
\caption{Average tokens in each mention for ChemNER+ and CHEMET datasets with few-shot settings.}
\label{img:length}
\end{figure}
\paragraph{Performance of Mention Extraction.} We calculate the mention F1 scores in Tables~\ref{tab:few_ChemNER+_mention} and \ref{tab:few_chemet_mention}. In addition, we also test a fully unsupervised mention extraction based on AMR-Parser~\cite{fernandez-astudillo-etal-2020-transition}\footnote{Implementation details are in Appendix~\ref{app:hyper}.}. The F1-scores are 38.22 and 45.33 for ChemNER+ and CHEMET, respectively. These results imply that the self-validation model generally improves the mention extraction accuracy. Moreover, adding decoder contrastive loss generally further bolsters the mention F1 score by reducing the number of tokens that appear in each mention, as shown in Figure~\ref{img:length}.

\begin{table}[!htb]
\centering
\small
\begin{tabularx}{\linewidth}{>{\hsize=2.15\hsize}X>{\centering\arraybackslash\hsize=0.77\hsize}X>{\centering\arraybackslash\hsize=0.77\hsize}X>{\centering\arraybackslash\hsize=0.77\hsize}X>{\centering\arraybackslash\hsize=0.77\hsize}X>{\centering\arraybackslash\hsize=0.77\hsize}X}
\toprule
 \textbf{$k$-shot} &\textbf{6} &\textbf{9}&\textbf{12} &\textbf{15}&\textbf{18} \\
 \midrule
RoBERTa              &  2.04 &  2.05&   2.05&   0.00& 2.05\\ 
PubMedBERT           &  2.05 &  0.00&   0.00&   2.13& 0.00\\\ 
ScholarBERT          &  0.00 &  9.28&   4.71&   0.00& 6.90\\ 
\hdashline
InBoXBART            &  8.33  & 11.36&  15.22& 17.14 & 7.69\\ 
+ Valid              & 10.81  & 12.24&  10.26& 9.76 & 23.81\\ 
+ Valid + CL         & \textbf{26.19}  & \textbf{23.91}&  \textbf{23.26}& \textbf{19.05} & \textbf{25.00}\\ 
\bottomrule
\end{tabularx}
\caption{micro-F1 (\%) scores for long-tail entity types ChemNER+ with few-shot settings. 
}
\label{tab:few_ChemNER+_long}
\end{table}

\begin{table}[!htb]

\centering
\small
\begin{tabularx}{\linewidth}{>{\hsize=2.15\hsize}X>{\centering\arraybackslash\hsize=0.77\hsize}X>{\centering\arraybackslash\hsize=0.77\hsize}X>{\centering\arraybackslash\hsize=0.77\hsize}X>{\centering\arraybackslash\hsize=0.77\hsize}X>{\centering\arraybackslash\hsize=0.77\hsize}X}
\toprule
 \textbf{$k$-shot} &\textbf{6} &\textbf{9}&\textbf{12} &\textbf{15}&\textbf{18} \\
 \midrule
RoBERTa              & 0.00  & 0.00& 0.00 & 0.00 & 0.00\\ 
PubMedBERT           & 0.00  & 0.00& 0.00 & 0.00 & 0.00\\ 
ScholarBERT          & 0.00  & 0.00& 0.00 & 0.00 & 0.00\\ 
\hdashline
InBoXBART            & 4.90  & 7.55& 4.55 & 5.05 & 12.26\\ 
+ Valid              & \textbf{8.72}  & \textbf{13.10}& 4.55 & \textbf{16.96} & 20.83\\ 
+ Valid + CL         & 7.07  & 11.32& \textbf{8.33} & 5.15 & \textbf{23.01}\\ 
\bottomrule
\end{tabularx}
\caption{micro-F1 (\%) scores for long-tail entity types CHEMET with few-shot settings. The encoder-based models fail to extract long-tail entity types for all few-shot settings. Compared to encoder-based models, seq2seq models can utilize label semantics in the generation procedure. Therefore, encoder-based models require more training data under few-shot settings.
}
\label{tab:few_chemet_long}
\end{table}

\paragraph{Performance of Long-tail Entity.} To evaluate the performance of long-tail entities, we first rank entity types by their frequency. We then select the entity types that appear in the lower 50\% and calculate the F1 scores of those types\footnote{Entity frequency and selected types are in Appendix~\ref{app:data}.}. The results are in Tables~\ref{tab:few_ChemNER+_long} and~\ref{tab:few_chemet_long}. Notably, our proposed methods greatly outperform the encoder-based baselines. Both the self-verification module and the decoder contrastive loss aid the entity extraction module in focusing on long-tail entities by creating a more balanced distribution of entity types. The major reason for the relatively low performance in Table~\ref{tab:few_chemet_long} is that the differences between the types in CHEMET are not significant. The relatively stable performance of our model in Table~\ref{tab:few_ChemNER+_long} across increasing few-shot examples indicates that our model achieves satisfactory performance for long-tail entities, even with a limited training sample.

%% file: 6analysis.tex
\section{Analysis}

\input{6.2qual}

\input{6.3cross}
\input{6.4error}

%% file: 6.2qual.tex
\subsection{Qualitative Analysis}

Table~\ref{tab:example} shows two typical examples from the 18-shot ChemNER+ dataset that illustrate how incorporating a self-validation module and decoder contrastive loss can improve the mention coverage and long-tail entity performance.

In the first example, the InBoXBART baseline fails to identify both \textit{``cyclophanes''} and \textit{``polycycles''}, probably because the input sentence contains too many entities. With the help of the self-validation module, the InBoXBART+Valid model successfully captures the first entity \textit{``cyclophanes''}. However, it still cannot recognize \textit{``polycycles''}. Additionally, both the baseline and the InBoXBART+Valid model mistakenly treat the entity \textit{``Suzuki cross-coupling and metathesis''} and the entity \textit{``metathesis''}, because those models excessively copy from the original sentence. In contrast, by adding the decoder contrastive loss, which uses the mentions with surrounding unrelated contexts as negatives, the model successfully separates the entity \textit{``Suzuki cross-coupling and metathesis''} from the entity \textit{``metathesis''}.

In the second example, both the baseline and the InBoXBART+Valid model predict a very long text span that treats three entities as a single entity. They also fail to capture \textit{``asymmetric catalysis''} and \textit{``highly enantioselective process''} as entities because their types have low frequency in the training set. With the help of decoder contrastive loss, the model reduces the excessive copying of the entity extraction module while trying to capture important entities as accurately as possible. Therefore, the model successfully classifies \textit{``asymmetric catalysis''} as \textit{Catalysis} correctly and also predicts \textit{``enantioselective process''} as an entity.

\begin{table*}[!htb]
\centering
\small
\renewcommand{\arraystretch}{1.4} 
\begin{tabularx}{\linewidth}{>{\arraybackslash\hsize=0.2\hsize}X>{\arraybackslash\hsize=1.8\hsize}X}
\toprule

 \textbf{InBoXBART}
 &
 Several \textcolor{myred}{\textit{cyclophanes}}, \textcolor{myred}{\textit{polycycles}}, ... have been synthesized by employing a combination of \textcolor{myorange}{\textit{Suzuki cross-coupling and metathesis}\textsubscript{ Coupling reactions}}.
\\
 \hdashline
 \textbf{+Valid}
& 
Several \textcolor{myorange}{\textit{cyclophanes}\textsubscript{ Heterocyclic compounds}}, \textcolor{myred}{\textit{polycycles}}, ... have been synthesized by employing a combination of  \textcolor{myorange}{\textit{Suzuki cross-coupling and metathesis}\textsubscript{ Organic reactions}}.
\\
 \hdashline
 \textbf{+Valid+CL}
& 
Several \textcolor{myorange}{\textit{cyclophanes}\textsubscript{ Heterocyclic compounds}}, \textcolor{myorange}{\textit{polycycles}\textsubscript{ Biomolecules}}, ... have been synthesized by employing a combination of \textcolor{mygreen}{Suzuki cross-coupling\textsubscript{ Coupling reactions}} and \textcolor{myorange}{\textit{metathesis}\textsubscript{ Chemical properties}}.
\\
 \hdashline
\textbf{Ground Truth} 
&
Several \textcolor{mygreen}{cyclophanes\textsubscript{ Aromatic compounds}}, \textcolor{mygreen}{polycycles\textsubscript{ Organic polymers}}, ... have been synthesized by employing a combination of \textcolor{mygreen}{Suzuki cross-coupling\textsubscript{ Coupling reactions}} and \textcolor{mygreen}{metathesis\textsubscript{ Substitution reactions}}.
\\

\midrule
 \textbf{InBoXBART}
& 
... with the advantages of \textcolor{myred}{\textit{asymmetric catalysis}} (step and atom economy) in a rare example of an \textcolor{myorange}{\textit{enantioselective cross coupling of a racemic electrophile bearing an oxygen leaving group}\textsubscript{ Catalysis}} ... the identification of a \textcolor{myred}{\textit{highly enantioselective process}}.\\
 \hdashline
 \textbf{+Valid}
& 
 ... with the advantages of \textcolor{myred}{\textit{asymmetric catalysis}} (step and atom economy) in a rare example of an \textcolor{myorange}{\textit{enantioselective cross coupling of a racemic electrophile bearing an oxygen leaving group}\textsubscript{ Organometallic compounds}} ... the identification of a \textcolor{myred}{\textit{highly enantioselective process}}
\\
 \hdashline
 \textbf{+Valid+CL}
& 
...with the advantages of \textcolor{mygreen}{asymmetric catalysis\textsubscript{ Catalysis}} (step and atom economy) in a rare example of an \textcolor{myred}{\textit{enantioselective cross coupling}} of a \textcolor{myred}{\textit{racemic electrophile}} bearing an \textcolor{mygreen}{oxygen leaving group\textsubscript{ Functional groups}} ... the identification of a highly \textcolor{myorange}{enantioselective process\textsubscript{ Chemical properties}}.
\\
 \hdashline
\textbf{Ground Truth}
&
... with the advantages of \textcolor{mygreen}{asymmetric catalysis\textsubscript{ Catalysis}} ( step and atom economy ) in a rare example of an \textcolor{mygreen}{enantioselective cross coupling\textsubscript{ Coupling reactions}} of a \textcolor{mygreen}{racemic electrophile\textsubscript{ Organic compounds}} bearing an \textcolor{mygreen}{oxygen leaving group\textsubscript{ Functional groups}} ... the identification of a \textcolor{mygreen}{highly enantioselective process\textsubscript{ Catalysis}}.
\\
\bottomrule
\end{tabularx}
\caption{
Examples showing how the self-validation module and entity decoder contrastive loss improves the model performance.
We highlight \textcolor{mygreen}{Complete Correct}, \textcolor{myred}{\textit{Missed Entity}}, and \textcolor{myorange}{\textit{Partially Correct Prediction}} with different color. Compared to other baselines, our  \textbf{+Valid+CL} successfully captures entities where other baselines miss. 
}
\label{tab:example}
\end{table*}

%% file: 6.3cross.tex
\subsection{Compatible with Other Few-shot Datasets?}

\paragraph{CrossNER Dataset.}
In the above experiments, we focus on the few-shot settings for chemical papers and prove the effectiveness of our proposed framework.
To evaluate the generalization ability of our proposed framework on other domains, we conduct experiments on the CrossNER dataset~\cite{liu2021crossner}. The detailed statistics are in Table~\ref{tab:stat_crossner}.
We remove sentences without any entity. 
Because the CrossNER dataset is based on Wikipedia articles, we choose RoBERTa and ScholarBERT as encoder-based baselines. Additionally, we select BART-base~\cite{lewis-etal-2020-bart} as the backbone for our ablation variations. 

 \begin{table}[!htb]
\centering
\small
\begin{tabularx}{\linewidth}{>{\hsize=0.8\hsize}X>{\centering\arraybackslash\hsize=0.9\hsize}X>{\centering\arraybackslash\hsize=0.8\hsize}X>{\centering\arraybackslash\hsize=0.8\hsize}X>{\centering\arraybackslash\hsize=0.8\hsize}X>{\centering\arraybackslash\hsize=1.4\hsize}X>{\centering\arraybackslash\hsize=1.4\hsize}X}
\toprule
\textbf{Dom.}&\textbf{Train}   & \textbf{Valid}& \textbf{Test} & \textbf{\#Type}& $\mathbf{\overline{\#Token}}$ & $\mathbf{\overline{\#Entity}}$  \\
\midrule
AI    & 100 & 350 & 430 & 14 & 31.5 & 4.42\\ 
Lit.  & 99  & 400 & 416 & 12 & 37.6 & 5.39\\ 
Mus.  & 100 & 380 & 465 & 13 & 41.4 & 7.05\\ 
Pol.  & 200 & 541 & 651 & 9  & 43.5 & 6.46\\ 
Sci.  & 200 & 450 & 543 & 17 & 35.8 & 5.62\\ 
\bottomrule
\end{tabularx}
\caption{Statistics of CrossNER. \textit{Dom.} denotes the domain of the dataset. \label{tab:stat_crossner} }
\end{table}

\begin{table}[!htb]

\centering
\small
\begin{tabularx}{\linewidth}{>{\hsize=2.15\hsize}X>{\centering\arraybackslash\hsize=0.77\hsize}X>{\centering\arraybackslash\hsize=0.77\hsize}X>{\centering\arraybackslash\hsize=0.77\hsize}X>{\centering\arraybackslash\hsize=0.77\hsize}X>{\centering\arraybackslash\hsize=0.77\hsize}X}
\toprule
 \textbf{Model} &\textbf{AI} &\textbf{Lit.}&\textbf{Mus.} &\textbf{Pol.}&\textbf{Sci.} \\
 \midrule
RoBERTa              &  60.88 & 67.51& 59.07 & 63.79 & 60.96\\ 
ScholarBERT          &  56.99 & 59.35& 52.26 & 57.15 & 57.01\\ 
\hdashline
BART-base            &  59.20 & 66.90& 62.78 & 67.99 & 62.18\\ 
+ Valid              &  61.84 & 67.97& 60.94 & 67.22 & 62.40\\ 
+ Valid + CL         &  \textbf{62.48} & \textbf{68.22}& \textbf{63.39} & \textbf{68.03} & \textbf{62.87}\\ 
\bottomrule
\end{tabularx}
\caption{F1 (\%) scores for CrossNER.}
\label{tab:crossner}
\end{table}

\paragraph{Results.}
As shown in Table~\ref{tab:crossner}, our model consistently produces the best F1 scores across all five domains of CrossNER without any external knowledge or domain adaptive pretraining. We observe that the model achieves the largest gain for the AI domain and the smallest gain for the politics domain. The major reason behind this is that AI domain contains the most informative entity types, which cover the key points of the sentence, including \textit{algorithm}, \textit{task}, etc. On the contrary, the politics domain contains many names of \textit{politicians} and \textit{locations}, which require background knowledge for the self-verification module to identify.


%% file: 6.4error.tex
\subsection{Remaining Challenges}
\paragraph{Misleading Subwords.} We observe that the mention text can sometimes mislead the type predictions, especially if the type contains a subword from the mention. As a result, the model fails to identify the type correctly. For example, given the mention \textit{``unnatural amino acid derivatives''}, our model focuses on the word ``acid'' and predicts the entity to be \textit{Organic acids} instead of \textit{Organonitrogen compounds}. The potential reason behind this is that the BART model incorrectly associates the \textit{``acid''} in the mention with \textit{Organic acids}. Such type errors might be incorporated into the decoder contrastive learning as additional hard negatives.

\paragraph{Fine-grained Type Classification.} The model tends to predict generic entity types instead of more fine-grained entity types. For instance, the model predicts the mention \textit{``Cs2CO3''} as \textit{Inorganic compounds} instead of \textit{Inorganic carbon compounds}. This issue might come from annotation ambiguity in the training set. Additionally, the model predicts types that are not in the predefined ontology. For instance, the model labels \textit{``GK''} as \textit{Genecyclic compounds} instead of \textit{Enzymes}. This error can possibly be solved by constraint decoding.

%% file: 7related.tex
\section{Related Work}
\paragraph{Scientific Entity Extraction.} Entity extraction for scientific papers has been widely exploited in the biomedical domain~\cite{nguyen-etal-2022-hardness,labrak-etal-2023-drbert,cao-etal-2023-gaussian,li-etal-2023-multi-source,hiebel-etal-2023-synthetic} and the computer science domain~\cite{luan-etal-2018-multi,jain-etal-2020-scirex,viswanathan-etal-2021-citationie,TriMF,ye-etal-2022-packed,SciDeBERTa,hong-etal-2023-diminishing}. Despite this, fine-grained scientific entity extraction~\cite{wang-etal-2021-chemner} in the chemical domain receives less attention due to the scarcity of benchmark resources. Most benchmarks in the chemical ~\cite{krallinger2015chemdner,kim2015overview} only provide coarse-grained entity types. In this paper, we address this problem by releasing two new datasets for chemical fine-grained entity extraction based on the ChemNER schema~\cite{wang-etal-2021-chemner} and CHEMET dataset~\cite{checnkai2021}. 

\paragraph{Few-shot Entity Extraction.} Few-shot learning attracts growing interest, especially for low-resource domains. Previous improvements for few-shot learning can be divided into several categories: domain-adaptive training by training the model in the same or similar domains~\cite{liu2021crossner,oh2022understanding}, prototype learning by learning entity type prototypes~\cite{ji-etal-2022-shot,oh2022understanding,ma-etal-2023-coarse}, prompt-based methods~\cite{lee-etal-2022-good,xu-etal-2023-focusing,nookala-etal-2023-adversarial,yang-etal-2023-mixpave,chen-etal-2023-prompt}, data-augmentation~\cite{cai-etal-2023-graph,ghosh-etal-2023-aclm}, code generation~\cite{li-etal-2023-codeie}, meta-learning~\cite{de-lichy-etal-2021-meta,metafew,ma-etal-2022-decomposed}, knowledge distillation~\cite{metaself,chen-etal-2023-learning}, contrastive learning~\cite{das-etal-2022-container}, and external knowledge including label definitions~\cite{wang-etal-2021-learning-language-description}, AMR graph~\cite{zhang-etal-2021-fine}, and background knowledge~\cite{lai-etal-2021-joint}. In contrast to these methods, our approach formulates the task in a text-to-text framework. In addition, we introduce a new simple but effective self-validation module, which achieves competitive performance without external knowledge or domain adaptive training.

\paragraph{Cycle Consistency.} Cycle consistency, namely structural duality, leverages the symmetric structure of tasks to facilitate the learning process. It has emerged as an effective way to deal with low-resource tasks in natural language processing. First introduced in machine translation~\cite{NIPS2016_5b69b9cb,cheng-etal-2016-semi,lample2018unsupervised,mohiuddin-joty-2019-revisiting,xu-etal-2020-dual} to deal with the scarcity of parallel data, cycle consistency has been expanded to other natural language processing tasks, including semantic parsing~\cite{cao-etal-2019-semantic,ye-etal-2019-jointly}, natural language understanding~\cite{su-etal-2019-dual,tseng-etal-2020-generative,su-etal-2020-towards}, and data-to-text generation~\cite{dognin-etal-2020-dualtkb,guo-etal-2020-cyclegt,wang-etal-2023-faithful}. Recently, \citet{cyclener} successfully apply the cycle consistency to entity extraction by introducing an iterative two-stage cycle consistency training procedure. Despite these efforts, the non-differentiability of the intermediate text in the cycle remains unsolved, leading to the inability to propagate the loss through the cycle. To address this issue, \citet{cyclener} and \citet{wang-etal-2023-faithful} alternatively freeze one of the two models in two adjacent cycles. On the contrary, we introduce the gumbel-softmax estimator to avoid the non-differentiable issue. Additionally, we reduce the dual cycle training into end-to-end training to save time and computation resources. 

%% file: 8conclusion.tex
\section{Conclusion and Future Work}
In this paper, we introduce a novel framework for chemical fine-grained entity extraction. Specifically, we target two unique challenges for few-shot fine-grained scientific entity extraction: mention coverage and long-tail entity extraction. We build a new self-validation module to automatically proofread the entity extraction results and a novel decoder contrastive loss to reduce excessive copying. Experimental results show that our proposed model achieves significant performance gains on two datasets: ChemNER+ and CHEMET. In the future, we plan to explore incorporating an external knowledge base to further improve the model's performance. Specifically, we plan to inject type definition into the representation to facilitate the entity extraction procedure. We will also continue exploring the use of constraint decoding to further improve entity extraction quality. 

%% file: 9limitations.tex
\section{Limitations}
\subsection{Limitations of Data Collections}
Both ChemNER+ and CHEMET are based on papers about Suzuki Coupling reactions from PubMed\footnote{\url{https://pubmed.ncbi.nlm.nih.gov/}}. Our fine-grained entity extraction datasets are biased towards the topics and ontology provided by ChemNER+ and CHEMET. For example, CHEMET only focuses on the organic compounds. The number of available sentences is limited by the original dataset and our annotation efforts. We currently only focus on the English sentences. We only test our model on chemical papers (i.e., ChemNER+ and CHEMET) and Wikipedia (CrossNER). In the future, we aim to adapt our model for categories in other languages.

\subsection{Limitations of System Performance} Our few-shot learning framework currently requires defining the entity ontology and few-shot examples before performing any training and testing. Therefore, due to patterns in the pretraining set, our model might produce mention types that don't align with our predefined ontology. For instance, it may generate \textit{Cyclopentadienyl compounds} instead of the predefined type \textit{Cyclopentadienyl complexes}.  Furthermore, the pretrained model might emphasize language modeling over accurately identifying entire chemical phrases. For example, it might recognize \textit{Pd} in the catalyst \textit{Pd(OAC)2} simply as a \textit{transition metal}.

%% file: appendix.tex
\input{App/1hyper}
\input{App/2data}
\input{App/3total_eval}

\input{App/4sci}
\input{App/5human}
\input{App/6ethical}

%% file: App/1hyper.tex
\section{Training and Evaluation Details}
\label{app:hyper}

\begin{table}[!htb]
\centering
\small
\begin{tabularx}{\linewidth}{>{\hsize=1\hsize}X>{\centering\arraybackslash\hsize=1\hsize}X>{\centering\arraybackslash\hsize=1\hsize}X}
\toprule
&\textbf{Avg. runtime}&\textbf{\# of Parameters}  \\
\midrule
RoBERTa                   &    16min       & 125M \\ 
PubMedBERT                &    18min       & 109M \\ 
ScholarBERT               &    19min       & 355M \\ 
InBoXBART                 &    58min       & 139M \\ 
\hspace{1mm}+Valid        &    56min       & 279M\\ 
\hspace{1mm}+Valid+CL     &    59min       & 279M\\ 
\bottomrule
\end{tabularx}
\caption{Runtimne (exclude CrossNER) and Number of Model Parameters \label{tab:para}}
\end{table}

Our baselines and model are based on the Huggingface framework~\cite{wolf-etal-2020-transformers}\footnote{\url{https://github.com/huggingface/transformers}}. Our models are trained on a single NVIDIA A100 GPU. All hyperparameter settings are listed below. We optimize all models by AdamW \cite{Loshchilov2019DecoupledWD}. The runtime and  number of parameters is listed in Table~\ref{tab:para}.

\paragraph{RoBERTa.} We train a \textit{RoBERTa-base}  model with 100 epochs and a batch size 32. The learning rate is $2\times 10 ^{-5}$ with $\epsilon=1\times 10 ^{-6}$. We use a linear scheduler for the optimizer. 

\paragraph{PubMedBERT.} The PubMedBERT has the same model architecture as \textit{BERT-base} with 12 transformer layers. The original checkpoint is pretrained on PubMed abstracts and full-text articles. We train a \textit{microsoft/BiomedNLP-PubMedBERT-base-uncased-abstract-fulltext} model with 100 epochs and a batch size 32. The learning rate is $2\times 10 ^{-5}$ with $\epsilon=1\times 10 ^{-6}$. We use a linear scheduler for the optimizer. 

\paragraph{ScholarBERT.} The ScholarBERT is based on the same architecture as \textit{BERT-large}. The original checkpoint is pretrained on 5,496,055 articles from 178,928 journals. The pretraining corpus has 45.3\% articles about biomedicine and life sciences. We train a \textit{globuslabs/ScholarBERT} model with 100 epochs and a batch size 32. The learning rate is $2\times 10 ^{-5}$ with $\epsilon=1\times 10 ^{-6}$. We use a linear scheduler for the optimizer.

\paragraph{InBoXBART.} The InBoXBART is an instructional-tuning language model for 32 biomedical NLP tasks based on \textit{BART-base}. We train the \textit{cogint/in-boxbart} model with 100 epochs and a batch size 16. The learning rate is $10 ^{-5}$ with $\epsilon=1\times 10 ^{-6}$. During decoding, we use beam-search to generate results with a beam size 5. We use cosine annealing warm restarts schedule \citep{DBLP:conf/iclr/LoshchilovH17} for the optimizer.

\paragraph{InBoXBART+Valid.} We first pretrain the self-validation model, which is based on \textit{cogint/in-boxbart}, on the training set. The learning rate for the self-validation module is $1\times 10 ^{-5}$ with $\epsilon=1\times 10 ^{-6}$. We use BLUE and ROUGE to select the best model. We then train the entity extraction model and the self-validation model jointly with cross-entropy $\mathcal{L}_{\mathrm{gen}}$ loss and reconstruction loss $\mathcal{L}_{\mathrm{recon}}$. The final loss is $  \mathcal{L} = \mathcal{L}_{\mathrm{gen}} + 5\cdot\mathcal{L}_{\mathrm{recon}}$. The learning rate is $5\times 10 ^{-5}$ with $\epsilon=1\times 10 ^{-6}$. During decoding, we use beam-search to generate results with a beam size 5. We use cosine annealing warm restarts schedule \citep{DBLP:conf/iclr/LoshchilovH17} for the optimizer.
 
\paragraph{InBoXBART+Valid+CL.} The final model is similar to \textit{InBoXBART+Valid}. We retain the self-validation module and add a new decoder contrastive loss. The final loss is $  \mathcal{L} = \mathcal{L}_{\mathrm{gen}}  + 0.2 \cdot \mathcal{L}_{\mathrm{cl}}+ 5\cdot\mathcal{L}_{\mathrm{recon}}$. We randomly choose 5 negative samples for each instance. The learning rate is $5\times 10 ^{-5}$ with $\epsilon=1\times 10 ^{-6}$. During decoding, we use beam-search to generate results with a beam size 5. We use cosine annealing warm restarts schedule \citep{DBLP:conf/iclr/LoshchilovH17} for the optimizer.

\paragraph{AMR-based Mention Extraction.} We use AMR-parser~\cite{fernandez-astudillo-etal-2020-transition} to extract mentions. We treat all text spans that are linkable to Wikipedia as mentions. 

\paragraph{NNShot and StructShot.} We use the implementation from \citet{ding-etal-2021-nerd} and choose \textit{RoBERTa-base} as the language model.

\paragraph{Evaluation Metrics.} We use entity-level micro-F1 for all experiments. We use the library from nereval
\url{https://github.com/jantrienes/nereval}. 

%% file: App/2data.tex
\section{Dataset Details}
\label{app:data}
We list the entity types of ChemNER+ and CHEMET below:
\begin{itemize}
    \item ChemNER+: Transition metals, Organic acids, Heterocyclic compounds, Organometallic compounds, Reagents for organic chemistry, Inorganic compounds, Thermodynamic properties, Aromatic compounds, Metal halides, Organic reactions, Alkylating agents, Organic compounds, Coupling reactions, Functional groups, Inorganic silicon compounds, Stereochemistry, Organohalides, Chemical properties, Catalysts, Free radicals, Alkaloids, Coordination chemistry, Ligands, Organophosphorus compounds, Reactive intermediates, Substitution reactions, Inorganic carbon compounds, Organonitrogen compounds, Biomolecules, Coordination compounds, Halogens, Chemical elements, Chlorides, Elimination reactions, Organic redox reactions, Inorganic phosphorus compounds, Organic polymers, Macrocycles, Cyclopentadienyl complexes, Substituents, Name reactions, Spiro compounds, Chemical kinetics, Organometallic chemistry, Catalysis, Organosulfur compounds, Ring forming reactions, Noble gases, Protecting groups, Addition reactions, Carbenes, Inorganic nitrogen compounds, Non-coordinating anions, Polymerization reactions, Carbon-carbon bond forming reactions, Isomerism, Enzymes, Oxoacids, Hydrogenation catalysts
    \item CHEMET: Acyl Groups, Alkanes, Alkenes, Alkynes, Amides, Amines, Aryl Groups, Carbenes, Carboxylic Acids, Esters, Ethers, Heterocyclic Compounds, Ketones, Nitriles, Nitro Compounds, Organic Polymers, Organohalides, Organometallic Compounds, Other Aromatic Compounds, Other Hydrocarbons, Other Organic Acids, Other Organic Compounds, Other Organonitrogen Compounds, Other Organophosphorus Compounds, Phosphinic Acids And Derivatives, Phosphonic Acids, Phosphonic Acids And Derivatives, Polycyclic Organic Compounds, Sulfonic Acids, Thiols
\end{itemize}

The frequency for each type in the training data of both ChemNER+ and CHEMET are listed below:
\begin{itemize}
    \item ChemNER+:
     Organic compounds: 183,
     Coupling reactions: 171,
     Aromatic compounds: 136,
     Functional groups: 120,
     Heterocyclic compounds: 106,
     Catalysts: 70,
     Biomolecules: 66,
     Chemical elements: 64,
     Organohalides: 63,
     Transition metals: 56,
     Chemical properties: 55,
     Ligands: 55,
     Organic acids: 48,
     Thermodynamic properties: 43,
     Inorganic compounds: 43,
     Coordination compounds: 37,
     Stereochemistry: 33,
     Organometallic compounds: 33,
     Reagents for organic chemistry: 28,
     Coordination chemistry: 27,
     Organonitrogen compounds: 26,
     Organic reactions: 23,
     Organic polymers: 23,
     Substitution reactions: 21,
     Catalysis: 20,
     Organic redox reactions: 18,
     Reactive intermediates: 13,
     Substituents: 13,
     Halogens: 12,
     Addition reactions: 8,
     Chlorides: 6,
     Ring forming reactions: 6,
     Inorganic carbon compounds: 6,
     Enzymes: 6,
     Alkaloids: 4,
     Organophosphorus compounds: 4,
     Organosulfur compounds: 4,
     Oxoacids: 4,
     Elimination reactions: 3,
     Carbenes: 3,
     Inorganic phosphorus compounds: 2,
     Chemical kinetics: 2,
     Macrocycles: 2,
     Noble gases: 2,
     Organometallic chemistry: 2,
     Hydrogenation catalysts: 2,
     Metal halides: 1,
     Cyclopentadienyl complexes: 1,
     Inorganic nitrogen compounds: 1,
     Protecting groups: 1,
     Alkylating agents: 1,
     Polymerization reactions: 1

    \item CHEMET:  Other Organic Compounds: 1705,
     Ethers: 934,
     Other Aromatic Compounds: 882,
     Heterocyclic Compounds: 792,
     Alkanes: 528,
     Amides: 516,
     Other Organonitrogen Compounds: 501,
     Organometallic Compounds: 495,
     Esters: 440,
     Amines: 431,
     Ketones: 406,
     Polycyclic Organic Compounds: 375,
     Aryl Groups: 363,
     Organohalides: 312,
     Alkynes: 281,
     Alkenes: 266,
     Organic Polymers: 255,
     Other Hydrocarbons: 236,
     Other Organic Acids: 194,
     Other Organophosphorus Compounds: 97,
     Acyl Groups: 78,
     Nitriles: 77,
     Carboxylic Acids: 62,
     Sulfonic Acids: 37,
     Nitro Compounds: 26,
     Carbenes: 9,
     Phosphonic Acids And Derivatives: 4,
     Thiols: 2
\end{itemize}

We consider the following types as long-tail entity types for ChemNER+ and CHEMET. We list both the entity type and its frequency:
\begin{itemize}
    \item ChemNER+:    Reactive intermediates: 13,
    Substituents: 13,
    Halogens: 12,
    Addition reactions: 8,
    Chlorides: 6,
    Ring forming reactions: 6,
    Inorganic carbon compounds: 6,
    Enzymes: 6,
    Alkaloids: 4,
    Organophosphorus compounds: 4,
    Organosulfur compounds: 4,
    Oxoacids: 4,
    Elimination reactions: 3,
    Carbenes: 3,
    Inorganic phosphorus compounds: 2,
    Chemical kinetics: 2,
    Macrocycles: 2,
    Noble gases: 2,
    Organometallic chemistry: 2,
    Hydrogenation catalysts: 2,
    Metal halides: 1,
    Cyclopentadienyl complexes: 1,
    Inorganic nitrogen compounds: 1,
    Protecting groups: 1,
    Alkylating agents: 1,
    Polymerization reactions: 1
    \item CHEMET:  Alkynes: 281,
    Alkenes: 266,
    Organic Polymers: 255,
    Other Hydrocarbons: 236,
    Other Organic Acids: 194,
    Other Organophosphorus Compounds: 97,
    Acyl Groups: 78,
    Nitriles: 77,
    Carboxylic Acids: 62,
    Sulfonic Acids: 37,
    Nitro Compounds: 26,
    Carbenes: 9,
    Phosphonic Acids And Derivatives: 4,
    Thiols: 2
\end{itemize}

%% file: App/3total_eval.tex
\section{Evaluation on Whole Dataset}

\begin{table}[!htb]

\centering
\small
\begin{tabularx}{\linewidth}{>{\hsize=1.75\hsize}X>{\centering\arraybackslash\hsize=.75\hsize}X>{\centering\arraybackslash\hsize=.75\hsize}X>{\centering\arraybackslash\hsize=.75\hsize}X}
\toprule
 \textbf{Model} &\textbf{Precision} &\textbf{Recall} &\textbf{F1} \\
 \midrule
In-BoXBART           & 55.73 &43.28   &48.72\\
+ Valid              & \textbf{57.49} &45.77   &50.97\\
+ Valid + CL         & 57.41 &\textbf{46.20}   &\textbf{51.10}\\

\bottomrule
\end{tabularx}
\caption{micro-F1 for ChemNER+ with the whole training set.
}
\label{tab:chemner}
\end{table}

We conduct fully supervised training on all training sets. The results are listed in Table~\ref{tab:chemner} and \ref{tab:CHEMET}. We observe that the self-validation module still improves the performance of the original InBoXBART for two datasets. We observe that the decoder contrastive loss further improves the model performance on ChemNER+. However, adding the entity decoder contrastive loss slightly decreases it. Because there are 6561 sentences in the CHEMET dataset, which is larger than the ChemNER+ dataset, the model with the self-validation module already performs very well. Additionally, since the CHEMET model contains fewer entities per sentence than the ChemNER+ dataset and these entities are all organic compounds separated away from each other, the entity decoder contrastive loss might introduce noise into the generation results, consequently decreasing the performance.

\begin{table}[!htb]

\centering
\small
\begin{tabularx}{\linewidth}{>{\hsize=1.75\hsize}X>{\centering\arraybackslash\hsize=.75\hsize}X>{\centering\arraybackslash\hsize=.75\hsize}X>{\centering\arraybackslash\hsize=.75\hsize}X}
\toprule
 \textbf{Model} &\textbf{Precision} &\textbf{Recall} &\textbf{F1} \\
 \midrule
In-BoXBART           & 64.94 &41.62   &50.73\\
+ Valid              & \textbf{70.09} &\textbf{42.16}   &\textbf{52.65}\\
+ Valid + CL         & 68.50 &41.31   &51.15  \\

\bottomrule
\end{tabularx}
\caption{micro-F1 for CHEMET with the whole training set.
}
\label{tab:CHEMET}
\end{table}

%% file: App/4sci.tex
\section{Scientific Artifacts}
We list the licenses of the scientific artifacts used in this paper: PMC Open Access Subset~\cite{gamble2017pubmed}\footnote{\url{https://www.ncbi.nlm.nih.gov/pmc/tools/openftlist/}} (CC BY-NC, CC BY-NC-SA, CC BY-NC-ND licenses), Huggingface Transformers (Apache License 2.0), ChemNER (no license), CHEMET\footnote{\url{https://github.com/chenkaisun/MMLI1}} (MIT license), RoBERTa (cc-by-4.0), PubMedBERT (MIT license), ScholarBERT (apache-2.0), BLEU\footnote{\url{https://github.com/cocodataset/cocoapi/blob/master/license.txt}}, ROUGE\footnote{\url{https://github.com/cocodataset/cocoapi/blob/master/license.txt}}, InBoXBART (MIT license), brat (MIT license), and nereval
 (MIT license). Our usage of existing artifacts is consistent with their intended use.

%% file: App/5human.tex
\section{Human Annotation}
\label{app:human}
The instructions for human annotations can be found in the supplementary material. Human annotators are required to annotate the chemical compound entities mentioned either in natural language or chemical formulas and other chemical related terms including reactions, catalysts, etc. We recruit two senior Ph.D. students from the Chemistry department in our university to perform human annotations. We use brat~\cite{stenetorp-etal-2012-brat} for all human annotations.

%% file: App/6ethical.tex
\section{Ethical Consideration}
The Chem-FINESE model and corresponding models we have designed in this paper are limited to the chemical domain, and might not be applicable to other scenarios.

\subsection{Usage Requirement}
Our Chem-FINESE system provides investigative leads for few-shot fine-grained entity extraction for the chemical domain. Therefore, the final results are not meant to be used without any human review. However, domain experts might be able to use this tool as a research assistant in scientific discovery. In addition, our system does not perform fact-checking or incorporate any external knowledge, which remains as future work. Our model is trained on PubMed papers written in English, which might present language barriers for readers who have been historically underrepresented in the NLP/Chemical domain.

\subsection{Data Collection}
Our ChemNER+ sentences are based on papers from PMC Open Access Subset. Our annotation is approved by the IRB at our university. All annotators involved in the human evaluation are voluntary participants and receive a fair wage. Our dataset can only be used for non-commercial purposes based on PMC Open Access Terms of Use.

%% file: acl_latex.bbl
\begin{thebibliography}{71}
\expandafter\ifx\csname natexlab\endcsname\relax\def\natexlab#1{#1}\fi

\bibitem[{Cai et~al.(2023)Cai, Huang, Jiang, Tan, Xie, and
  Tu}]{cai-etal-2023-graph}
Jiong Cai, Shen Huang, Yong Jiang, Zeqi Tan, Pengjun Xie, and Kewei Tu. 2023.
\newblock \href {https://doi.org/10.18653/v1/2023.acl-short.11} {Graph
  propagation based data augmentation for named entity recognition}.
\newblock In \emph{Proceedings of the 61st Annual Meeting of the Association
  for Computational Linguistics (Volume 2: Short Papers)}, pages 110--118,
  Toronto, Canada. Association for Computational Linguistics.

\bibitem[{Cao et~al.(2023)Cao, Peek, Renehan, and
  Ananiadou}]{cao-etal-2023-gaussian}
Jiarun Cao, Niels Peek, Andrew Renehan, and Sophia Ananiadou. 2023.
\newblock \href {https://doi.org/10.18653/v1/2023.bionlp-1.2} {{G}aussian
  distributed prototypical network for few-shot genomic variant detection}.
\newblock In \emph{The 22nd Workshop on Biomedical Natural Language Processing
  and BioNLP Shared Tasks}, pages 26--36, Toronto, Canada. Association for
  Computational Linguistics.

\bibitem[{Cao et~al.(2019)Cao, Zhu, Liu, Li, and Yu}]{cao-etal-2019-semantic}
Ruisheng Cao, Su~Zhu, Chen Liu, Jieyu Li, and Kai Yu. 2019.
\newblock \href {https://doi.org/10.18653/v1/P19-1007} {Semantic parsing with
  dual learning}.
\newblock In \emph{Proceedings of the 57th Annual Meeting of the Association
  for Computational Linguistics}, pages 51--64, Florence, Italy. Association
  for Computational Linguistics.

\bibitem[{Chen et~al.(2023{\natexlab{a}})Chen, Lu, Lin, Lou, Jia, Dai, Wu, Cao,
  Han, and Sun}]{chen-etal-2023-learning}
Jiawei Chen, Yaojie Lu, Hongyu Lin, Jie Lou, Wei Jia, Dai Dai, Hua Wu, Boxi
  Cao, Xianpei Han, and Le~Sun. 2023{\natexlab{a}}.
\newblock \href {https://doi.org/10.18653/v1/2023.acl-long.764} {Learning
  in-context learning for named entity recognition}.
\newblock In \emph{Proceedings of the 61st Annual Meeting of the Association
  for Computational Linguistics (Volume 1: Long Papers)}, pages 13661--13675,
  Toronto, Canada. Association for Computational Linguistics.

\bibitem[{Chen et~al.(2023{\natexlab{b}})Chen, Zheng, and
  Yang}]{chen-etal-2023-prompt}
Yanru Chen, Yanan Zheng, and Zhilin Yang. 2023{\natexlab{b}}.
\newblock \href {https://doi.org/10.18653/v1/2023.findings-acl.451}
  {Prompt-based metric learning for few-shot {NER}}.
\newblock In \emph{Findings of the Association for Computational Linguistics:
  ACL 2023}, pages 7199--7212, Toronto, Canada. Association for Computational
  Linguistics.

\bibitem[{Cheng et~al.(2016)Cheng, Xu, He, He, Wu, Sun, and
  Liu}]{cheng-etal-2016-semi}
Yong Cheng, Wei Xu, Zhongjun He, Wei He, Hua Wu, Maosong Sun, and Yang Liu.
  2016.
\newblock \href {https://doi.org/10.18653/v1/P16-1185} {Semi-supervised
  learning for neural machine translation}.
\newblock In \emph{Proceedings of the 54th Annual Meeting of the Association
  for Computational Linguistics (Volume 1: Long Papers)}, pages 1965--1974,
  Berlin, Germany. Association for Computational Linguistics.

\bibitem[{Das et~al.(2022)Das, Katiyar, Passonneau, and
  Zhang}]{das-etal-2022-container}
Sarkar Snigdha~Sarathi Das, Arzoo Katiyar, Rebecca Passonneau, and Rui Zhang.
  2022.
\newblock \href {https://doi.org/10.18653/v1/2022.acl-long.439} {{CONT}ai{NER}:
  Few-shot named entity recognition via contrastive learning}.
\newblock In \emph{Proceedings of the 60th Annual Meeting of the Association
  for Computational Linguistics (Volume 1: Long Papers)}, pages 6338--6353,
  Dublin, Ireland. Association for Computational Linguistics.

\bibitem[{de~Lichy et~al.(2021)de~Lichy, Glaude, and
  Campbell}]{de-lichy-etal-2021-meta}
Cyprien de~Lichy, Hadrien Glaude, and William Campbell. 2021.
\newblock \href {https://doi.org/10.18653/v1/2021.metanlp-1.6} {Meta-learning
  for few-shot named entity recognition}.
\newblock In \emph{Proceedings of the 1st Workshop on Meta Learning and Its
  Applications to Natural Language Processing}, pages 44--58, Online.
  Association for Computational Linguistics.

\bibitem[{Ding et~al.(2021)Ding, Xu, Chen, Wang, Han, Xie, Zheng, and
  Liu}]{ding-etal-2021-nerd}
Ning Ding, Guangwei Xu, Yulin Chen, Xiaobin Wang, Xu~Han, Pengjun Xie, Haitao
  Zheng, and Zhiyuan Liu. 2021.
\newblock \href {https://doi.org/10.18653/v1/2021.acl-long.248} {Few-{NERD}: A
  few-shot named entity recognition dataset}.
\newblock In \emph{Proceedings of the 59th Annual Meeting of the Association
  for Computational Linguistics and the 11th International Joint Conference on
  Natural Language Processing (Volume 1: Long Papers)}, pages 3198--3213,
  Online. Association for Computational Linguistics.

\bibitem[{Dognin et~al.(2020)Dognin, Melnyk, Padhi, Nogueira~dos Santos, and
  Das}]{dognin-etal-2020-dualtkb}
Pierre Dognin, Igor Melnyk, Inkit Padhi, Cicero Nogueira~dos Santos, and Payel
  Das. 2020.
\newblock \href {https://doi.org/10.18653/v1/2020.emnlp-main.694} {{D}ual{TKB}:
  {A} {D}ual {L}earning {B}ridge between {T}ext and {K}nowledge {B}ase}.
\newblock In \emph{Proceedings of the 2020 Conference on Empirical Methods in
  Natural Language Processing (EMNLP)}, pages 8605--8616, Online. Association
  for Computational Linguistics.

\bibitem[{Fernandez~Astudillo et~al.(2020)Fernandez~Astudillo, Ballesteros,
  Naseem, Blodgett, and Florian}]{fernandez-astudillo-etal-2020-transition}
Ram{\'o}n Fernandez~Astudillo, Miguel Ballesteros, Tahira Naseem, Austin
  Blodgett, and Radu Florian. 2020.
\newblock \href {https://doi.org/10.18653/v1/2020.findings-emnlp.89}
  {Transition-based parsing with stack-transformers}.
\newblock In \emph{Findings of the Association for Computational Linguistics:
  EMNLP 2020}, pages 1001--1007, Online. Association for Computational
  Linguistics.

\bibitem[{Gamble(2017)}]{gamble2017pubmed}
Alyson Gamble. 2017.
\newblock Pubmed central (pmc).
\newblock \emph{The Charleston Advisor}, 19(2):48--54.

\bibitem[{Ghosh et~al.(2023)Ghosh, Tyagi, Suri, Kumar, S, and
  Manocha}]{ghosh-etal-2023-aclm}
Sreyan Ghosh, Utkarsh Tyagi, Manan Suri, Sonal Kumar, Ramaneswaran S, and
  Dinesh Manocha. 2023.
\newblock \href {https://doi.org/10.18653/v1/2023.acl-long.8} {{ACLM}: A
  selective-denoising based generative data augmentation approach for
  low-resource complex {NER}}.
\newblock In \emph{Proceedings of the 61st Annual Meeting of the Association
  for Computational Linguistics (Volume 1: Long Papers)}, pages 104--125,
  Toronto, Canada. Association for Computational Linguistics.

\bibitem[{Giorgi et~al.(2022)Giorgi, Bader, and
  Wang}]{giorgi-etal-2022-sequence}
John Giorgi, Gary Bader, and Bo~Wang. 2022.
\newblock \href {https://doi.org/10.18653/v1/2022.bionlp-1.2} {A
  sequence-to-sequence approach for document-level relation extraction}.
\newblock In \emph{Proceedings of the 21st Workshop on Biomedical Language
  Processing}, pages 10--25, Dublin, Ireland. Association for Computational
  Linguistics.

\bibitem[{Gu et~al.(2021)Gu, Tinn, Cheng, Lucas, Usuyama, Liu, Naumann, Gao,
  and Poon}]{pubmedbert}
Yu~Gu, Robert Tinn, Hao Cheng, Michael Lucas, Naoto Usuyama, Xiaodong Liu,
  Tristan Naumann, Jianfeng Gao, and Hoifung Poon. 2021.
\newblock \href {https://doi.org/10.1145/3458754} {Domain-specific language
  model pretraining for biomedical natural language processing}.
\newblock \emph{ACM Trans. Comput. Healthcare}, 3(1).

\bibitem[{Guo et~al.(2020)Guo, Jin, Qiu, Zhang, Wipf, and
  Zhang}]{guo-etal-2020-cyclegt}
Qipeng Guo, Zhijing Jin, Xipeng Qiu, Weinan Zhang, David Wipf, and Zheng Zhang.
  2020.
\newblock \href {https://aclanthology.org/2020.webnlg-1.8} {{C}ycle{GT}:
  Unsupervised graph-to-text and text-to-graph generation via cycle training}.
\newblock In \emph{Proceedings of the 3rd International Workshop on Natural
  Language Generation from the Semantic Web (WebNLG+)}, pages 77--88, Dublin,
  Ireland (Virtual). Association for Computational Linguistics.

\bibitem[{He et~al.(2016)He, Xia, Qin, Wang, Yu, Liu, and
  Ma}]{NIPS2016_5b69b9cb}
Di~He, Yingce Xia, Tao Qin, Liwei Wang, Nenghai Yu, Tie-Yan Liu, and Wei-Ying
  Ma. 2016.
\newblock \href
  {https://proceedings.neurips.cc/paper_files/paper/2016/file/5b69b9cb83065d403869739ae7f0995e-Paper.pdf}
  {Dual learning for machine translation}.
\newblock In \emph{Advances in Neural Information Processing Systems},
  volume~29. Curran Associates, Inc.

\bibitem[{Hiebel et~al.(2023)Hiebel, Ferret, Fort, and
  N{\'e}v{\'e}ol}]{hiebel-etal-2023-synthetic}
Nicolas Hiebel, Olivier Ferret, Karen Fort, and Aur{\'e}lie N{\'e}v{\'e}ol.
  2023.
\newblock \href {https://doi.org/10.18653/v1/2023.eacl-main.170} {Can synthetic
  text help clinical named entity recognition? a study of electronic health
  records in {F}rench}.
\newblock In \emph{Proceedings of the 17th Conference of the European Chapter
  of the Association for Computational Linguistics}, pages 2320--2338,
  Dubrovnik, Croatia. Association for Computational Linguistics.

\bibitem[{Hong et~al.(2023)Hong, Ajith, Pauloski, Duede, Chard, and
  Foster}]{hong-etal-2023-diminishing}
Zhi Hong, Aswathy Ajith, James Pauloski, Eamon Duede, Kyle Chard, and Ian
  Foster. 2023.
\newblock \href {https://doi.org/10.18653/v1/2023.findings-acl.82} {The
  diminishing returns of masked language models to science}.
\newblock In \emph{Findings of the Association for Computational Linguistics:
  ACL 2023}, pages 1270--1283, Toronto, Canada. Association for Computational
  Linguistics.

\bibitem[{Iovine et~al.(2022)Iovine, Fang, Fetahu, Rokhlenko, and
  Malmasi}]{cyclener}
Andrea Iovine, Anjie Fang, Besnik Fetahu, Oleg Rokhlenko, and Shervin Malmasi.
  2022.
\newblock \href {https://doi.org/10.1145/3485447.3512012} {Cyclener: An
  unsupervised training approach for named entity recognition}.
\newblock In \emph{Proceedings of the ACM Web Conference 2022}, WWW '22, page
  2916–2924, New York, NY, USA. Association for Computing Machinery.

\bibitem[{Jain et~al.(2020)Jain, van Zuylen, Hajishirzi, and
  Beltagy}]{jain-etal-2020-scirex}
Sarthak Jain, Madeleine van Zuylen, Hannaneh Hajishirzi, and Iz~Beltagy. 2020.
\newblock \href {https://doi.org/10.18653/v1/2020.acl-main.670} {{S}ci{REX}:
  {A} challenge dataset for document-level information extraction}.
\newblock In \emph{Proceedings of the 58th Annual Meeting of the Association
  for Computational Linguistics}, pages 7506--7516, Online. Association for
  Computational Linguistics.

\bibitem[{Jang et~al.(2017)Jang, Gu, and Poole}]{jang2017categorical}
Eric Jang, Shixiang Gu, and Ben Poole. 2017.
\newblock \href {https://openreview.net/forum?id=rkE3y85ee} {Categorical
  reparameterization with gumbel-softmax}.
\newblock In \emph{Poceedings of 5th International Conference on Learning
  Representations}.

\bibitem[{Jeong and Kim(2022)}]{SciDeBERTa}
Yuna Jeong and Eunhui Kim. 2022.
\newblock \href {https://doi.org/10.1109/ACCESS.2022.3180830} {Scideberta:
  Learning deberta for science technology documents and fine-tuning information
  extraction tasks}.
\newblock \emph{IEEE Access}, 10:60805--60813.

\bibitem[{Ji et~al.(2022)Ji, Li, Gan, Yu, Ma, Liu, and
  Yang}]{ji-etal-2022-shot}
Bin Ji, Shasha Li, Shaoduo Gan, Jie Yu, Jun Ma, Huijun Liu, and Jing Yang.
  2022.
\newblock \href {https://aclanthology.org/2022.coling-1.159} {Few-shot named
  entity recognition with entity-level prototypical network enhanced by
  dispersedly distributed prototypes}.
\newblock In \emph{Proceedings of the 29th International Conference on
  Computational Linguistics}, pages 1842--1854, Gyeongju, Republic of Korea.
  International Committee on Computational Linguistics.

\bibitem[{Kim et~al.(2015)Kim, Dogan, Chatr-Aryamontri, Tyers, Wilbur, and
  Comeau}]{kim2015overview}
Sun Kim, Rezarta~Islamaj Dogan, Andrew Chatr-Aryamontri, Mike Tyers, W~John
  Wilbur, and Donald~C Comeau. 2015.
\newblock \href
  {https://biocreative.bioinformatics.udel.edu/media/store/files/2015/BCV2015_BioC.pdf}
  {Overview of biocreative v bioc track}.
\newblock In \emph{Proceedings of the Fifth BioCreative Challenge Evaluation
  Workshop, Sevilla, Spain}, pages 1--9.

\bibitem[{Krallinger et~al.(2015)Krallinger, Rabal, Leitner, Vazquez, Salgado,
  Lu, Leaman, Lu, Ji, Lowe et~al.}]{krallinger2015chemdner}
Martin Krallinger, Obdulia Rabal, Florian Leitner, Miguel Vazquez, David
  Salgado, Zhiyong Lu, Robert Leaman, Yanan Lu, Donghong Ji, Daniel~M Lowe,
  et~al. 2015.
\newblock \href {https://doi.org/10.1186/1758-2946-7-S1-S2} {The chemdner
  corpus of chemicals and drugs and its annotation principles}.
\newblock \emph{Journal of cheminformatics}, 7(1):1--17.

\bibitem[{Labrak et~al.(2023)Labrak, Bazoge, Dufour, Rouvier, Morin, Daille,
  and Gourraud}]{labrak-etal-2023-drbert}
Yanis Labrak, Adrien Bazoge, Richard Dufour, Mickael Rouvier, Emmanuel Morin,
  B{\'e}atrice Daille, and Pierre-Antoine Gourraud. 2023.
\newblock \href {https://doi.org/10.18653/v1/2023.acl-long.896} {{D}r{BERT}: A
  robust pre-trained model in {F}rench for biomedical and clinical domains}.
\newblock In \emph{Proceedings of the 61st Annual Meeting of the Association
  for Computational Linguistics (Volume 1: Long Papers)}, pages 16207--16221,
  Toronto, Canada. Association for Computational Linguistics.

\bibitem[{Lai et~al.(2021)Lai, Ji, Zhai, and Tran}]{lai-etal-2021-joint}
Tuan Lai, Heng Ji, ChengXiang Zhai, and Quan~Hung Tran. 2021.
\newblock \href {https://doi.org/10.18653/v1/2021.acl-long.488} {Joint
  biomedical entity and relation extraction with knowledge-enhanced collective
  inference}.
\newblock In \emph{Proceedings of the 59th Annual Meeting of the Association
  for Computational Linguistics and the 11th International Joint Conference on
  Natural Language Processing (Volume 1: Long Papers)}, pages 6248--6260,
  Online. Association for Computational Linguistics.

\bibitem[{Lample et~al.(2018)Lample, Conneau, Denoyer, and
  Ranzato}]{lample2018unsupervised}
Guillaume Lample, Alexis Conneau, Ludovic Denoyer, and Marc'Aurelio Ranzato.
  2018.
\newblock \href {https://openreview.net/forum?id=rkYTTf-AZ} {Unsupervised
  machine translation using monolingual corpora only}.
\newblock In \emph{the Sixth International Conference on Learning
  Representations}.

\bibitem[{Landhuis(2016)}]{landhuis2016scientific}
Esther Landhuis. 2016.
\newblock \href {https://www.nature.com/articles/nj7612-457a} {Scientific
  literature: Information overload}.
\newblock \emph{Nature}, 535(7612):457--458.

\bibitem[{Lee et~al.(2022)Lee, Kadakia, Tan, Agarwal, Feng, Shibuya, Mitani,
  Sekiya, Pujara, and Ren}]{lee-etal-2022-good}
Dong-Ho Lee, Akshen Kadakia, Kangmin Tan, Mahak Agarwal, Xinyu Feng, Takashi
  Shibuya, Ryosuke Mitani, Toshiyuki Sekiya, Jay Pujara, and Xiang Ren. 2022.
\newblock \href {https://doi.org/10.18653/v1/2022.acl-long.192} {Good examples
  make a faster learner: Simple demonstration-based learning for low-resource
  {NER}}.
\newblock In \emph{Proceedings of the 60th Annual Meeting of the Association
  for Computational Linguistics (Volume 1: Long Papers)}, pages 2687--2700,
  Dublin, Ireland. Association for Computational Linguistics.

\bibitem[{Lewis et~al.(2020)Lewis, Liu, Goyal, Ghazvininejad, Mohamed, Levy,
  Stoyanov, and Zettlemoyer}]{lewis-etal-2020-bart}
Mike Lewis, Yinhan Liu, Naman Goyal, Marjan Ghazvininejad, Abdelrahman Mohamed,
  Omer Levy, Veselin Stoyanov, and Luke Zettlemoyer. 2020.
\newblock \href {https://doi.org/10.18653/v1/2020.acl-main.703} {{BART}:
  Denoising sequence-to-sequence pre-training for natural language generation,
  translation, and comprehension}.
\newblock In \emph{Proceedings of the 58th Annual Meeting of the Association
  for Computational Linguistics}, pages 7871--7880, Online. Association for
  Computational Linguistics.

\bibitem[{Li et~al.(2022)Li, Chiu, Feng, and Wang}]{metafew}
Jing Li, Billy Chiu, Shanshan Feng, and Hao Wang. 2022.
\newblock \href {https://doi.org/10.1109/TKDE.2020.3038670} {Few-shot named
  entity recognition via meta-learning}.
\newblock \emph{IEEE Transactions on Knowledge and Data Engineering},
  34(9):4245--4256.

\bibitem[{Li et~al.(2023{\natexlab{a}})Li, Sun, Tang, Yan, Wu, Huang, and
  Qiu}]{li-etal-2023-codeie}
Peng Li, Tianxiang Sun, Qiong Tang, Hang Yan, Yuanbin Wu, Xuanjing Huang, and
  Xipeng Qiu. 2023{\natexlab{a}}.
\newblock \href {https://doi.org/10.18653/v1/2023.acl-long.855} {{C}ode{IE}:
  Large code generation models are better few-shot information extractors}.
\newblock In \emph{Proceedings of the 61st Annual Meeting of the Association
  for Computational Linguistics (Volume 1: Long Papers)}, pages 15339--15353,
  Toronto, Canada. Association for Computational Linguistics.

\bibitem[{Li et~al.(2023{\natexlab{b}})Li, Martschat, and
  Ponzetto}]{li-etal-2023-multi-source}
Yueling Li, Sebastian Martschat, and Simone~Paolo Ponzetto. 2023{\natexlab{b}}.
\newblock \href {https://doi.org/10.18653/v1/2023.bionlp-1.1} {Multi-source
  (pre-)training for cross-domain measurement, unit and context extraction}.
\newblock In \emph{The 22nd Workshop on Biomedical Natural Language Processing
  and BioNLP Shared Tasks}, pages 1--25, Toronto, Canada. Association for
  Computational Linguistics.

\bibitem[{Liu et~al.(2019)Liu, Ott, Goyal, Du, Joshi, Chen, Levy, Lewis,
  Zettlemoyer, and Stoyanov}]{liu2019roberta}
Yinhan Liu, Myle Ott, Naman Goyal, Jingfei Du, Mandar Joshi, Danqi Chen, Omer
  Levy, Mike Lewis, Luke Zettlemoyer, and Veselin Stoyanov. 2019.
\newblock \href {http://arxiv.org/abs/1907.11692} {Roberta: A robustly
  optimized bert pretraining approach}.
\newblock \emph{Computation and Language Repository}, arXiv:1907.11692.

\bibitem[{Liu et~al.(2021)Liu, Xu, Yu, Dai, Ji, Cahyawijaya, Madotto, and
  Fung}]{liu2021crossner}
Zihan Liu, Yan Xu, Tiezheng Yu, Wenliang Dai, Ziwei Ji, Samuel Cahyawijaya,
  Andrea Madotto, and Pascale Fung. 2021.
\newblock \href {https://arxiv.org/pdf/2012.04373.pdf} {Crossner: Evaluating
  cross-domain named entity recognition}.
\newblock In \emph{Proceedings of the AAAI Conference on Artificial
  Intelligence}, volume~35, pages 13452--13460.

\bibitem[{Loshchilov and Hutter(2017)}]{DBLP:conf/iclr/LoshchilovH17}
Ilya Loshchilov and Frank Hutter. 2017.
\newblock \href {https://openreview.net/forum?id=Skq89Scxx} {{SGDR:} stochastic
  gradient descent with warm restarts}.
\newblock In \emph{5th International Conference on Learning Representations,
  {ICLR} 2017, Toulon, France, April 24-26, 2017, Conference Track
  Proceedings}. OpenReview.net.

\bibitem[{Loshchilov and Hutter(2019)}]{Loshchilov2019DecoupledWD}
Ilya Loshchilov and Frank Hutter. 2019.
\newblock \href {https://openreview.net/pdf?id=Bkg6RiCqY7} {Decoupled weight
  decay regularization}.
\newblock In \emph{Proceedings of the 7th International Conference on Learning
  Representations}.

\bibitem[{Luan et~al.(2018)Luan, He, Ostendorf, and
  Hajishirzi}]{luan-etal-2018-multi}
Yi~Luan, Luheng He, Mari Ostendorf, and Hannaneh Hajishirzi. 2018.
\newblock \href {https://doi.org/10.18653/v1/D18-1360} {Multi-task
  identification of entities, relations, and coreference for scientific
  knowledge graph construction}.
\newblock In \emph{Proceedings of the 2018 Conference on Empirical Methods in
  Natural Language Processing}, pages 3219--3232, Brussels, Belgium.
  Association for Computational Linguistics.

\bibitem[{Ma et~al.(2023)Ma, Lin, Chen, Zhou, Wang, Gui, Zhang, Gao, and
  Chen}]{ma-etal-2023-coarse}
Ruotian Ma, Zhang Lin, Xuanting Chen, Xin Zhou, Junzhe Wang, Tao Gui, Qi~Zhang,
  Xiang Gao, and Yun~Wen Chen. 2023.
\newblock \href {https://doi.org/10.18653/v1/2023.findings-acl.253}
  {Coarse-to-fine few-shot learning for named entity recognition}.
\newblock In \emph{Findings of the Association for Computational Linguistics:
  ACL 2023}, pages 4115--4129, Toronto, Canada. Association for Computational
  Linguistics.

\bibitem[{Ma et~al.(2022)Ma, Jiang, Wu, Zhao, and
  Lin}]{ma-etal-2022-decomposed}
Tingting Ma, Huiqiang Jiang, Qianhui Wu, Tiejun Zhao, and Chin-Yew Lin. 2022.
\newblock \href {https://doi.org/10.18653/v1/2022.findings-acl.124} {Decomposed
  meta-learning for few-shot named entity recognition}.
\newblock In \emph{Findings of the Association for Computational Linguistics:
  ACL 2022}, pages 1584--1596, Dublin, Ireland. Association for Computational
  Linguistics.

\bibitem[{Mohiuddin and Joty(2019)}]{mohiuddin-joty-2019-revisiting}
Tasnim Mohiuddin and Shafiq Joty. 2019.
\newblock \href {https://doi.org/10.18653/v1/N19-1386} {Revisiting adversarial
  autoencoder for unsupervised word translation with cycle consistency and
  improved training}.
\newblock In \emph{Proceedings of the 2019 Conference of the North {A}merican
  Chapter of the Association for Computational Linguistics: Human Language
  Technologies, Volume 1 (Long and Short Papers)}, pages 3857--3867,
  Minneapolis, Minnesota. Association for Computational Linguistics.

\bibitem[{Nguyen et~al.(2022)Nguyen, Du, Buntine, Chen, and
  Beare}]{nguyen-etal-2022-hardness}
Ngoc~Dang Nguyen, Lan Du, Wray Buntine, Changyou Chen, and Richard Beare. 2022.
\newblock \href {https://doi.org/10.18653/v1/2022.emnlp-main.271}
  {Hardness-guided domain adaptation to recognise biomedical named entities
  under low-resource scenarios}.
\newblock In \emph{Proceedings of the 2022 Conference on Empirical Methods in
  Natural Language Processing}, pages 4063--4071, Abu Dhabi, United Arab
  Emirates. Association for Computational Linguistics.

\bibitem[{Nookala et~al.(2023)Nookala, Verma, Mukherjee, and
  Kumar}]{nookala-etal-2023-adversarial}
Venkata Prabhakara~Sarath Nookala, Gaurav Verma, Subhabrata Mukherjee, and
  Srijan Kumar. 2023.
\newblock \href {https://doi.org/10.18653/v1/2023.findings-acl.138}
  {Adversarial robustness of prompt-based few-shot learning for natural
  language understanding}.
\newblock In \emph{Findings of the Association for Computational Linguistics:
  ACL 2023}, pages 2196--2208, Toronto, Canada. Association for Computational
  Linguistics.

\bibitem[{Oh et~al.(2022)Oh, Kim, Ho, Kim, Song, and Yun}]{oh2022understanding}
Jaehoon Oh, Sungnyun Kim, Namgyu Ho, Jin-Hwa Kim, Hwanjun Song, and Se-Young
  Yun. 2022.
\newblock \href {https://openreview.net/forum?id=rH-X09cB50f} {Understanding
  cross-domain few-shot learning based on domain similarity and few-shot
  difficulty}.
\newblock In \emph{Advances in Neural Information Processing Systems}.

\bibitem[{Oord et~al.(2018)Oord, Li, and Vinyals}]{oord2018representation}
Aaron van~den Oord, Yazhe Li, and Oriol Vinyals. 2018.
\newblock \href {http://arxiv.org/abs/1807.03748} {Representation learning with
  contrastive predictive coding}.
\newblock \emph{Machine Learning Repository}, arXiv:1807.03748.

\bibitem[{Parmar et~al.(2022)Parmar, Mishra, Purohit, Luo, Mohammad, and
  Baral}]{parmar-etal-2022-boxbart}
Mihir Parmar, Swaroop Mishra, Mirali Purohit, Man Luo, Murad Mohammad, and
  Chitta Baral. 2022.
\newblock \href {https://doi.org/10.18653/v1/2022.findings-naacl.10}
  {In-{B}o{XBART}: Get instructions into biomedical multi-task learning}.
\newblock In \emph{Findings of the Association for Computational Linguistics:
  NAACL 2022}, pages 112--128, Seattle, United States. Association for
  Computational Linguistics.

\bibitem[{Shen et~al.(2021)Shen, Ma, Tang, and Lu}]{TriMF}
Yongliang Shen, Xinyin Ma, Yechun Tang, and Weiming Lu. 2021.
\newblock \href {https://doi.org/10.1145/3442381.3449895} {A trigger-sense
  memory flow framework for joint entity and relation extraction}.
\newblock In \emph{Proceedings of the Web Conference 2021}, WWW '21, page
  1704–1715, New York, NY, USA. Association for Computing Machinery.

\bibitem[{Stenetorp et~al.(2012)Stenetorp, Pyysalo, Topi{\'c}, Ohta, Ananiadou,
  and Tsujii}]{stenetorp-etal-2012-brat}
Pontus Stenetorp, Sampo Pyysalo, Goran Topi{\'c}, Tomoko Ohta, Sophia
  Ananiadou, and Jun{'}ichi Tsujii. 2012.
\newblock \href {https://aclanthology.org/E12-2021} {brat: a web-based tool for
  {NLP}-assisted text annotation}.
\newblock In \emph{Proceedings of the Demonstrations at the 13th Conference of
  the {E}uropean Chapter of the Association for Computational Linguistics},
  pages 102--107, Avignon, France. Association for Computational Linguistics.

\bibitem[{Su et~al.(2019)Su, Huang, and Chen}]{su-etal-2019-dual}
Shang-Yu Su, Chao-Wei Huang, and Yun-Nung Chen. 2019.
\newblock \href {https://doi.org/10.18653/v1/P19-1545} {Dual supervised
  learning for natural language understanding and generation}.
\newblock In \emph{Proceedings of the 57th Annual Meeting of the Association
  for Computational Linguistics}, pages 5472--5477, Florence, Italy.
  Association for Computational Linguistics.

\bibitem[{Su et~al.(2020)Su, Huang, and Chen}]{su-etal-2020-towards}
Shang-Yu Su, Chao-Wei Huang, and Yun-Nung Chen. 2020.
\newblock \href {https://doi.org/10.18653/v1/2020.acl-main.63} {Towards
  unsupervised language understanding and generation by joint dual learning}.
\newblock In \emph{Proceedings of the 58th Annual Meeting of the Association
  for Computational Linguistics}, pages 671--680, Online. Association for
  Computational Linguistics.

\bibitem[{Sun et~al.(2021)Sun, Li, Xiao, Parulian, Zhai, and Ji}]{checnkai2021}
C.~Sun, W.~Li, J.~Xiao, N.~Parulian, C.~Zhai, and H.~Ji. 2021.
\newblock \href {https://doi.org/10.1109/BIBM52615.2021.9669360} {Fine-grained
  chemical entity typing with multimodal knowledge representation}.
\newblock In \emph{2021 IEEE International Conference on Bioinformatics and
  Biomedicine (BIBM)}, pages 1984--1991, Los Alamitos, CA, USA. IEEE Computer
  Society.

\bibitem[{Tjong Kim~Sang and
  De~Meulder(2003)}]{tjong-kim-sang-de-meulder-2003-introduction}
Erik~F. Tjong Kim~Sang and Fien De~Meulder. 2003.
\newblock \href {https://aclanthology.org/W03-0419} {Introduction to the
  {C}o{NLL}-2003 shared task: Language-independent named entity recognition}.
\newblock In \emph{Proceedings of the Seventh Conference on Natural Language
  Learning at {HLT}-{NAACL} 2003}, pages 142--147.

\bibitem[{Tseng et~al.(2020)Tseng, Cheng, Fang, and
  Vandyke}]{tseng-etal-2020-generative}
Bo-Hsiang Tseng, Jianpeng Cheng, Yimai Fang, and David Vandyke. 2020.
\newblock \href {https://doi.org/10.18653/v1/2020.acl-main.163} {A generative
  model for joint natural language understanding and generation}.
\newblock In \emph{Proceedings of the 58th Annual Meeting of the Association
  for Computational Linguistics}, pages 1795--1807, Online. Association for
  Computational Linguistics.

\bibitem[{Van~Noorden(2014)}]{van2014global}
Richard Van~Noorden. 2014.
\newblock \href
  {http://www.as.utexas.edu/astronomy/education/fall15/wheeler/secure/scientific_output_9.pdf}
  {Global scientific output doubles every nine years}.
\newblock \emph{Nature news blog}.

\bibitem[{Viswanathan et~al.(2021)Viswanathan, Neubig, and
  Liu}]{viswanathan-etal-2021-citationie}
Vijay Viswanathan, Graham Neubig, and Pengfei Liu. 2021.
\newblock \href {https://doi.org/10.18653/v1/2021.acl-long.59} {{C}itation{IE}:
  Leveraging the citation graph for scientific information extraction}.
\newblock In \emph{Proceedings of the 59th Annual Meeting of the Association
  for Computational Linguistics and the 11th International Joint Conference on
  Natural Language Processing (Volume 1: Long Papers)}, pages 719--731, Online.
  Association for Computational Linguistics.

\bibitem[{Wang et~al.(2023{\natexlab{a}})Wang, Li, Chan, Huang, Hockenmaier,
  Chowdhary, and Ji}]{wang-etal-2023-multimedia}
Qingyun Wang, Manling Li, Hou~Pong Chan, Lifu Huang, Julia Hockenmaier, Girish
  Chowdhary, and Heng Ji. 2023{\natexlab{a}}.
\newblock \href {https://doi.org/10.18653/v1/2023.findings-acl.63} {Multimedia
  generative script learning for task planning}.
\newblock In \emph{Findings of the Association for Computational Linguistics:
  ACL 2023}, pages 986--1008, Toronto, Canada. Association for Computational
  Linguistics.

\bibitem[{Wang et~al.(2021{\natexlab{a}})Wang, Hu, Song, Garg, Xiao, and
  Han}]{wang-etal-2021-chemner}
Xuan Wang, Vivian Hu, Xiangchen Song, Shweta Garg, Jinfeng Xiao, and Jiawei
  Han. 2021{\natexlab{a}}.
\newblock \href {https://doi.org/10.18653/v1/2021.emnlp-main.424} {{C}hem{NER}:
  Fine-grained chemistry named entity recognition with ontology-guided distant
  supervision}.
\newblock In \emph{Proceedings of the 2021 Conference on Empirical Methods in
  Natural Language Processing}, pages 5227--5240, Online and Punta Cana,
  Dominican Republic. Association for Computational Linguistics.

\bibitem[{Wang et~al.(2021{\natexlab{b}})Wang, Chu, Zhang, and
  Gao}]{wang-etal-2021-learning-language-description}
Yaqing Wang, Haoda Chu, Chao Zhang, and Jing Gao. 2021{\natexlab{b}}.
\newblock \href {https://doi.org/10.18653/v1/2021.findings-emnlp.139} {Learning
  from language description: Low-shot named entity recognition via decomposed
  framework}.
\newblock In \emph{Findings of the Association for Computational Linguistics:
  EMNLP 2021}, pages 1618--1630, Punta Cana, Dominican Republic. Association
  for Computational Linguistics.

\bibitem[{Wang et~al.(2021{\natexlab{c}})Wang, Mukherjee, Chu, Tu, Wu, Gao, and
  Awadallah}]{metaself}
Yaqing Wang, Subhabrata Mukherjee, Haoda Chu, Yuancheng Tu, Ming Wu, Jing Gao,
  and Ahmed~Hassan Awadallah. 2021{\natexlab{c}}.
\newblock \href {https://doi.org/10.1145/3447548.3467235} {Meta self-training
  for few-shot neural sequence labeling}.
\newblock In \emph{Proceedings of the 27th ACM SIGKDD Conference on Knowledge
  Discovery \& Data Mining}, KDD '21, page 1737–1747, New York, NY, USA.
  Association for Computing Machinery.

\bibitem[{Wang et~al.(2023{\natexlab{b}})Wang, Collins, Vedula, Filice,
  Malmasi, and Rokhlenko}]{wang-etal-2023-faithful}
Zhuoer Wang, Marcus Collins, Nikhita Vedula, Simone Filice, Shervin Malmasi,
  and Oleg Rokhlenko. 2023{\natexlab{b}}.
\newblock \href {https://doi.org/10.18653/v1/2023.acl-long.160} {Faithful
  low-resource data-to-text generation through cycle training}.
\newblock In \emph{Proceedings of the 61st Annual Meeting of the Association
  for Computational Linguistics (Volume 1: Long Papers)}, pages 2847--2867,
  Toronto, Canada. Association for Computational Linguistics.

\bibitem[{Wolf et~al.(2020)Wolf, Debut, Sanh, Chaumond, Delangue, Moi, Cistac,
  Rault, Louf, Funtowicz, Davison, Shleifer, von Platen, Ma, Jernite, Plu, Xu,
  Le~Scao, Gugger, Drame, Lhoest, and Rush}]{wolf-etal-2020-transformers}
Thomas Wolf, Lysandre Debut, Victor Sanh, Julien Chaumond, Clement Delangue,
  Anthony Moi, Pierric Cistac, Tim Rault, Remi Louf, Morgan Funtowicz, Joe
  Davison, Sam Shleifer, Patrick von Platen, Clara Ma, Yacine Jernite, Julien
  Plu, Canwen Xu, Teven Le~Scao, Sylvain Gugger, Mariama Drame, Quentin Lhoest,
  and Alexander Rush. 2020.
\newblock \href {https://doi.org/10.18653/v1/2020.emnlp-demos.6} {Transformers:
  State-of-the-art natural language processing}.
\newblock In \emph{Proceedings of the 2020 Conference on Empirical Methods in
  Natural Language Processing: System Demonstrations}, pages 38--45, Online.
  Association for Computational Linguistics.

\bibitem[{Xu et~al.(2020)Xu, Niu, and Carpuat}]{xu-etal-2020-dual}
Weijia Xu, Xing Niu, and Marine Carpuat. 2020.
\newblock \href {https://doi.org/10.18653/v1/2020.findings-emnlp.182} {Dual
  reconstruction: a unifying objective for semi-supervised neural machine
  translation}.
\newblock In \emph{Findings of the Association for Computational Linguistics:
  EMNLP 2020}, pages 2006--2020, Online. Association for Computational
  Linguistics.

\bibitem[{Xu et~al.(2023)Xu, Yang, Zhang, Zhou, Wu, and
  Zhou}]{xu-etal-2023-focusing}
Yuanyuan Xu, Zeng Yang, Linhai Zhang, Deyu Zhou, Tiandeng Wu, and Rong Zhou.
  2023.
\newblock \href {https://doi.org/10.18653/v1/2023.findings-acl.164} {Focusing,
  bridging and prompting for few-shot nested named entity recognition}.
\newblock In \emph{Findings of the Association for Computational Linguistics:
  ACL 2023}, pages 2621--2637, Toronto, Canada. Association for Computational
  Linguistics.

\bibitem[{Yan et~al.(2021)Yan, Gui, Dai, Guo, Zhang, and
  Qiu}]{yan-etal-2021-unified-generative}
Hang Yan, Tao Gui, Junqi Dai, Qipeng Guo, Zheng Zhang, and Xipeng Qiu. 2021.
\newblock \href {https://doi.org/10.18653/v1/2021.acl-long.451} {A unified
  generative framework for various {NER} subtasks}.
\newblock In \emph{Proceedings of the 59th Annual Meeting of the Association
  for Computational Linguistics and the 11th International Joint Conference on
  Natural Language Processing (Volume 1: Long Papers)}, pages 5808--5822,
  Online. Association for Computational Linguistics.

\bibitem[{Yang et~al.(2023)Yang, Wang, Wang, Quan, Feng, Chen, Khabsa, Wang,
  Xu, and Liu}]{yang-etal-2023-mixpave}
Li~Yang, Qifan Wang, Jingang Wang, Xiaojun Quan, Fuli Feng, Yu~Chen, Madian
  Khabsa, Sinong Wang, Zenglin Xu, and Dongfang Liu. 2023.
\newblock \href {https://doi.org/10.18653/v1/2023.findings-acl.633}
  {{M}ix{PAVE}: Mix-prompt tuning for few-shot product attribute value
  extraction}.
\newblock In \emph{Findings of the Association for Computational Linguistics:
  ACL 2023}, pages 9978--9991, Toronto, Canada. Association for Computational
  Linguistics.

\bibitem[{Yang and Katiyar(2020)}]{yang-katiyar-2020-simple}
Yi~Yang and Arzoo Katiyar. 2020.
\newblock \href {https://doi.org/10.18653/v1/2020.emnlp-main.516} {Simple and
  effective few-shot named entity recognition with structured nearest neighbor
  learning}.
\newblock In \emph{Proceedings of the 2020 Conference on Empirical Methods in
  Natural Language Processing (EMNLP)}, pages 6365--6375, Online. Association
  for Computational Linguistics.

\bibitem[{Ye et~al.(2022)Ye, Lin, Li, and Sun}]{ye-etal-2022-packed}
Deming Ye, Yankai Lin, Peng Li, and Maosong Sun. 2022.
\newblock \href {https://doi.org/10.18653/v1/2022.acl-long.337} {Packed
  levitated marker for entity and relation extraction}.
\newblock In \emph{Proceedings of the 60th Annual Meeting of the Association
  for Computational Linguistics (Volume 1: Long Papers)}, pages 4904--4917,
  Dublin, Ireland. Association for Computational Linguistics.

\bibitem[{Ye et~al.(2019)Ye, Li, and Wang}]{ye-etal-2019-jointly}
Hai Ye, Wenjie Li, and Lu~Wang. 2019.
\newblock \href {https://doi.org/10.18653/v1/P19-1201} {Jointly learning
  semantic parser and natural language generator via dual information
  maximization}.
\newblock In \emph{Proceedings of the 57th Annual Meeting of the Association
  for Computational Linguistics}, pages 2090--2101, Florence, Italy.
  Association for Computational Linguistics.

\bibitem[{Zhang et~al.(2021)Zhang, Parulian, Ji, Elsayed, Myers, and
  Palmer}]{zhang-etal-2021-fine}
Zixuan Zhang, Nikolaus Parulian, Heng Ji, Ahmed Elsayed, Skatje Myers, and
  Martha Palmer. 2021.
\newblock \href {https://doi.org/10.18653/v1/2021.acl-long.489} {Fine-grained
  information extraction from biomedical literature based on knowledge-enriched
  {A}bstract {M}eaning {R}epresentation}.
\newblock In \emph{Proceedings of the 59th Annual Meeting of the Association
  for Computational Linguistics and the 11th International Joint Conference on
  Natural Language Processing (Volume 1: Long Papers)}, pages 6261--6270,
  Online. Association for Computational Linguistics.

\end{thebibliography}
